\documentclass[11pt]{article}

\usepackage[preprint]{acl}

\usepackage{times}
\usepackage{latexsym}
\usepackage[T1]{fontenc}
\usepackage[utf8]{inputenc}
\usepackage{microtype}
\usepackage{inconsolata}
\usepackage{graphicx}
\usepackage{booktabs}
\usepackage{colortbl}   
\definecolor{cAnchored}{HTML}{dfe6ed}    
\definecolor{cRedundant}{HTML}{f5e3df}   
\definecolor{cBest}{HTML}{fef3e0}        
\newcommand{\hlanchored}[1]{\cellcolor{cAnchored}#1}
\newcommand{\hlredundant}[1]{\cellcolor{cRedundant}#1}
\newcommand{\hlbest}[1]{\cellcolor{cBest}\textbf{#1}}
\usepackage{amsmath}
\usepackage{amssymb}
\usepackage{subcaption}
\usepackage{multirow}
\usepackage{xurl}
\usepackage{wrapfig}
\usepackage{xcolor}
\usepackage{hyperref}

\definecolor{cLink}{HTML}{3d5a80}   
\definecolor{cUrl}{HTML}{2a8f82}    
\hypersetup{pdftitle={Are Single-Token Sparse Autoencoder Features Causally Necessary? Layer-Depth and SAE-Family Effects},
  pdfauthor={Seonglae Cho, Zekun Wu, Kleyton Da Costa, Rishi Kalra, Ilham Wicaksono, Adriano Koshiyama},
  colorlinks=true, citecolor=cLink, linkcolor=cLink, urlcolor=cUrl}

\usepackage{amsmath,amsfonts,bm}

\def\eqref#1{equation~\ref{#1}}

\def\1{\bm{1}}

\def\vw{{\bm{w}}}

\DeclareMathAlphabet{\mathsfit}{\encodingdefault}{\sfdefault}{m}{sl}
\SetMathAlphabet{\mathsfit}{bold}{\encodingdefault}{\sfdefault}{bx}{n}

\newcommand{\R}{\mathbb{R}}

\title{Are Single-Token Sparse Autoencoder Features Causally Necessary? \\ Layer-Depth and SAE-Family Effects}

\author{%
  Seonglae Cho\thanks{Correspondence: \texttt{seonglae.cho@holisticai.com}}\,$^{1,2}$ \quad
  Zekun Wu\,$^{1,2}$ \quad
  Kleyton Da Costa\,$^{1}$ \\[4pt]
  \textbf{Rishi Kalra}\,$^{1}$ \quad
  \textbf{Ilham Wicaksono}\,$^{1}$ \quad
  \textbf{Adriano Koshiyama}\,$^{1,2}$ \\[4pt]
  $^{1}$Holistic AI \quad $^{2}$University College London \\
}

\begin{document}

\maketitle

\begin{abstract}
\begingroup\exhyphenpenalty=10000
Sparse autoencoder (SAE) features are used to interpret and steer large language models, yet nobody has tested whether a feature's causal role is stable across SAE families.
Single-token features fire on one vocabulary item, so ground truth permits direct comparison.
We analyze 3.9M features across six models and three SAE families and zero-ablate at full layer depth: they sit 4.7$\times$ tighter in decoder space and concentrate in early layers.
Deleting one lowers the model's logit for that token in 178 of 208 layer conditions, significant after multiple-comparison correction.
And depth decides how the damage lands: early-layer deletions disrupt the layers that follow, late-layer deletions change the output directly.
Cross-family causal differences exceed within-family scale effects: on the same base model, GemmaScope and BatchTopK features are \emph{causally anchored}, LlamaScope features \emph{locally redundant}.
Under LlamaScope the token returns to within 2$\times$ its pre-ablation rank \mbox{96--98\%} of the time.
Changing only the activation function reverses the sign of that difference, so the training recipe is the remaining candidate: cross-family claims are sensitive to training methodology, not just activation function or scale.
\par\endgroup
\end{abstract}

\section{Introduction}
\label{sec:intro}

\begin{figure*}[t]
    \centering
    \includegraphics[width=\textwidth]{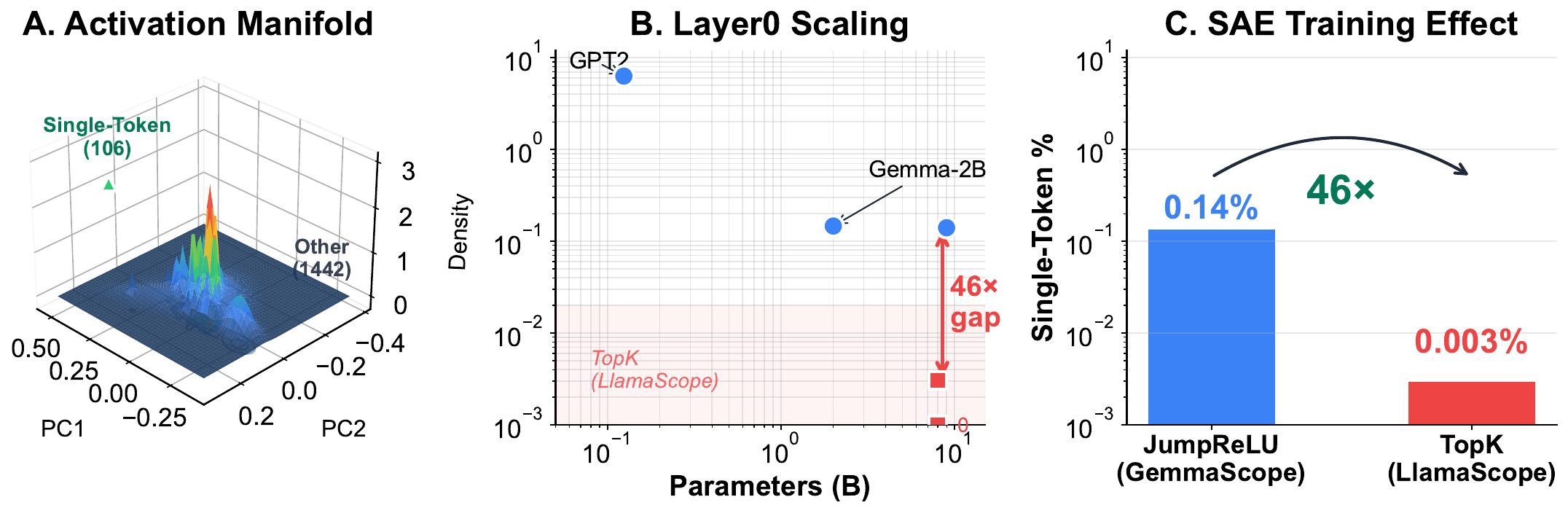}
    \caption{\textbf{Single-token prevalence and SAE family effects.}
    (A) PCA density surface of GPT2-Small Layer-0 activation vectors: single-token features form a sharp, concentrated peak while other features spread broadly (axes: top-2 principal components; height: kernel density).
    (B) Prevalence vs model scale: GemmaScope/res-jb (blue) declines 1.48\%$\to$0.14\%; LlamaScope (red) near-zero.
    No fit line ($n=3$).
    (C) GemmaScope shows 46$\times$ higher prevalence than LlamaScope at the 8--9B scale (Gemma-2-9B vs Llama-3.1-8B).}
    \label{fig:scaling_law}
\end{figure*}

\begingroup\exhyphenpenalty=10000
On the same base model, a feature whose ablation destroys the target token under one sparse autoencoder (SAE) family can be replaced almost immediately under another, leaving cross-family interpretability claims contingent on which SAE was applied.
We study \textbf{single-token features}, those activating on one vocabulary item and its morphological variants, as the SAE analog of grandmother cells \citep{gross2002genealogy}: the diagnostic endpoint of the monosemantic-polysemantic spectrum where vocabulary-level ground truth allows direct cross-family comparison.

And three reasons motivate this focus: (i) single-token features are the \emph{diagnostic case} with vocabulary-level ground truth, enabling direct validation; (ii) they form a measurable \emph{bridge} between embedding space and feature space, with higher embedding alignment than polysemantic features; (iii) they have \emph{practical relevance} for token-level reliability tasks where steering and editing must operate on token-identity directions.

We analyze 3.9M SAE features from Neuronpedia \citep{neuronpedia2024} across six models: GPT2-Small (124M) \citep{radford2019language, bloom2024gpt2residualsaes}, Gemma-2-2B and Gemma-2-9B \citep{gemmateam2024gemma2, lieberum2024gemma}, Gemma-3-1B \citep{gemmateam2025gemma3}, Llama-3.1-8B \citep{dubey2024llama}, and DeepSeek-R1-8B \citep{deepseekai2025deepseekr1}.
These span three SAE families: GemmaScope/res-jb, LlamaScope, and community BatchTopK.
The set spans nearly two orders of magnitude in scale.

Our analysis yields two main findings.
First, we find single-token features geometrically and causally distinct: 91\% of GPT2-Small's sit in Layer~0, and they cluster 4.7$\times$ tighter in decoder space.
The L0$\rightarrow$L1 transition is sharp, with Grassmannian alignment 0.26 there versus $>$0.9 for later layers: these features also persist more strongly across layers than polysemantic features.
Causal ablation across the seven full-depth model$\times$SAE configurations confirms necessity under zero-ablation at the measured readouts: 178 of their 208 layers are BH-significant.
Second, cross-SAE-family differences exceed within-family scale effects.
And LlamaScope SAEs show 46$\times$ lower prevalence than GemmaScope at comparable 8--9B scale, and token-matched comparisons on the same base model extend this gap to causal structure: GemmaScope and BatchTopK features are causally anchored while LlamaScope features are locally redundant, recovering their pre-ablation rank 96--98\% of the time, though activation function alone does not explain this (Section~\ref{sec:discussion}).
We additionally observe that causal importance varies by layer depth and semantic category.
Single-token features are the endpoint where ground truth is unambiguous and cross-family matching is exact.
If causal roles diverge at this simplest matched case, comparisons of more complex features, where establishing correspondence is itself contested, plausibly inherit at least this much instability, a scope argument rather than a demonstrated bound.
The takeaway: a feature's causal necessity is real but cannot be assumed portable across SAE families; treat the SAE family as an experimental variable to be checked, not a detail to be abstracted away.

\endgroup

\section{Related Work}
\label{sec:related}

\paragraph{SAE interpretability foundations.}
Sparse autoencoders decompose neural activations into interpretable features under the superposition hypothesis \citep{elhage2022superposition, bricken2023monosemanticity, huben2024sparse, shu2025survey}.
Monosemantic features at scale range from named entities to syntax patterns and abstract concepts \citep{templeton2024scaling, gao2025topksae}.
SAE features feed downstream steering, circuit analysis, and model editing pipelines \citep{marks2025sparse, chalnev2024steering, arad2025saes}, making cross-family stability practically consequential.

\paragraph{Cross-method evaluation and feature universality.}
The features SAEs recover depend on architectural choices and the metrics used to evaluate them \citep{leask2025canonical, locatello2019challenging, karvonen2025saebench, korznikov2026sanity}.
Recent work tracks feature evolution across layers \citep{balcells2024evolution, balagansky2025mechanistic, olah2020zoom, elhage2021mathematical} and tests cross-model universality \citep{lan2024universal, paulo2025sparseautoencoders, chanin2024absorption}.

\section{Methods}
\label{sec:methods}

\begin{figure*}[t]
    \centering
    \includegraphics[width=\textwidth]{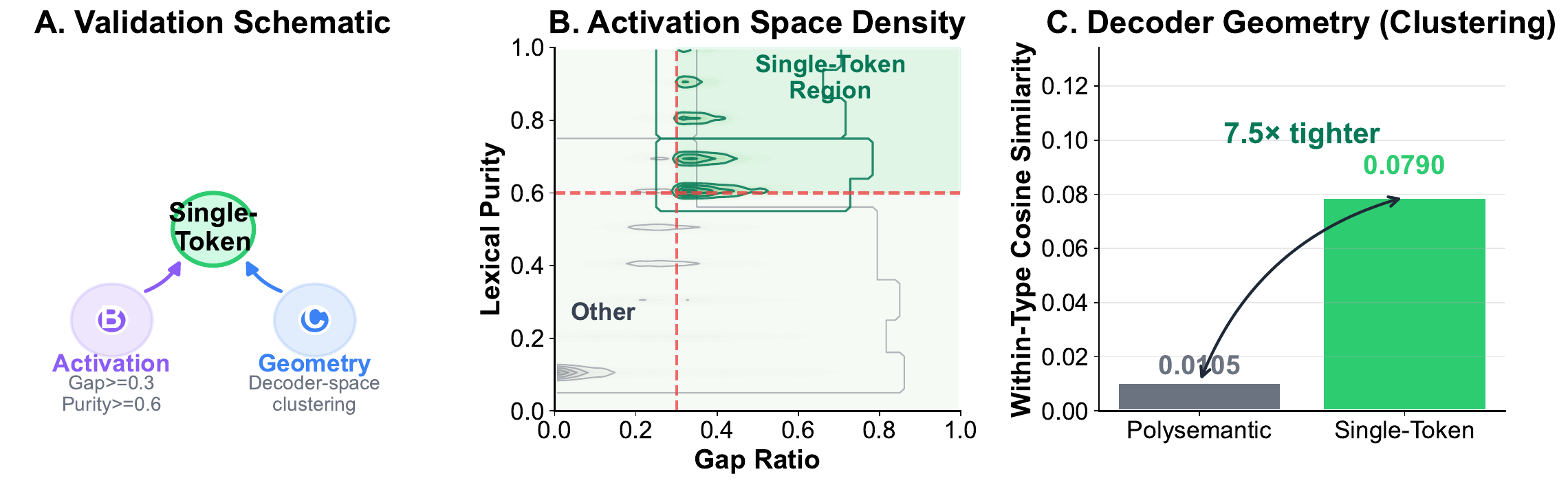}
    \caption{\textbf{Three-way validation.} (A) Detection schematic: activation metrics and decoder geometry. (B) Gap ratio vs lexical purity separates single-token (green) from polysemantic (gray). (C) Decoder-geometry clustering on GPT2-Small Layer~0: single-token decoder vectors are 7.5$\times$ tighter in mean pairwise cosine than a strict polysemantic set (gap $<0.2$, purity $<0.4$) read at top-10. At the canonical operating point used throughout the paper (top-20, full polysemantic complement) the ratio is 4.7$\times$ (Table~\ref{tab:thresholds}).}
    \label{fig:validation}
\end{figure*}

\subsection{Data and Models}

We analyze pre-trained SAEs hosted on Neuronpedia \citep{neuronpedia2024} across six models spanning three SAE families (Table~\ref{tab:models}).
All SAE checkpoints, activation data, and feature explanations are publicly available.
For each feature, we use: (i) decoder vectors $\vw_{\text{dec}} \in \R^{d}$ from SAE checkpoints \citep{chanind2024saelens}, (ii) top-$k$ activating tokens and values, and (iii) auto-interpretability explanations.
GPT2-Small uses res-jb SAEs \citep{bloom2024gpt2residualsaes}.
Gemma-2 and Gemma-3 models use GemmaScope JumpReLU SAEs \citep{lieberum2024gemma, rajamanoharan2024jumprelu, rajamanoharan2024gated}.
Llama-3.1 and DeepSeek-R1 use LlamaScope TopK SAEs \citep{he2024llamascope, gao2025topksae}, following the $k$-sparse autoencoder framework \citep{makhzani2013ksparse}.
For causal experiments, we additionally use community BatchTopK SAEs \citep{bussmann2024batchtopk, chanind2024saelens} on Gemma-2-2B and Gemma-3-1B to compare competitive versus independent activation functions on the same base models.

\begin{table}[t]
\centering
\caption{Models and SAE configurations analyzed.
SAE Family reflects training methodology.}
\label{tab:models}
\vspace{-2mm}
\small
\resizebox{\columnwidth}{!}{%
\begin{tabular}{@{}lccccccc@{}}
\toprule
\textbf{Model} & \textbf{Params} & \textbf{$d_{\text{model}}$} & \textbf{Layers} & \textbf{Feat/L} & \textbf{SAE Type} & \textbf{Family} & \textbf{Total} \\
\midrule
GPT2-Small & 124M & 768 & 12 & 24,576 & ReLU & res-jb & 294,912 \\
Gemma-2-2B & 2B & 2,304 & 26 & 16,384 & JumpReLU & GemmaScope & 425,984 \\
Gemma-2-9B & 9B & 3,584 & 42 & 16,384 & JumpReLU & GemmaScope & 688,128 \\
Gemma-3-1B & 1B & 1,152 & 26 & 16,384 & JumpReLU & GemmaScope & 425,984 \\
Llama-3.1-8B & 8B & 4,096 & 32 & 32,768 & TopK & LlamaScope & 1,048,576 \\
DeepSeek-R1 & 8B & 4,096 & 32 & 32,768 & TopK & LlamaScope & 1,048,576 \\
\bottomrule
\end{tabular}}
\end{table}

\subsection{Single-Token Detection}

We define a \textbf{single-token feature} as one whose activation is dominated by a single vocabulary item and its morphological variants.
We operationalize this concept using three continuous metrics computed from the top-$k=20$ activating tokens:

\textbf{Gap Ratio} measures how sharply the top token dominates:
\begin{equation}
\text{gap} = (v_1 - v_2)/v_1,
\label{eq:gap}
\end{equation}
where $v_1 \geq v_2$ are the top two activation values among the top-$k$ activating tokens.

\textbf{Lexical Purity} measures the fraction of the top-$k$ slots occupied by the most frequent token and its case variants:
\begin{equation}
\text{purity} = \max_t |\{i \in [k] : t_i = t\}|/k,
\label{eq:purity}
\end{equation}
where $t_i$ is the surface string of the $i$-th top activating token, normalized by stripping the word-boundary prefix and case-folding, and the maximum is taken over all unique normalized tokens $t$ in the top-$k$ set. Purity groups case and word-boundary variants of the same surface form; the activation-based implementation additionally groups inflectional variants sharing a stem, which is why Table~\ref{tab:examples} lists inflected forms within a single feature.

\textbf{Complete Word} requires a word-boundary prefix (space for byte pair encoding (BPE), U+2581 for SentencePiece), excluding subword fragments.

We select an operating point of gap $\geq 0.3$, purity $\geq 0.6$, and complete word (bolded row in Table~\ref{tab:thresholds}), validated by three independent signals: 64\% explanation match, 4.7$\times$ geometric clustering, and causal necessity under ablation.
This operating point is conservative: under a Dirichlet null over 20 tokens, the 99th-percentile gap ratio is 0.77, placing our threshold well below the null tail; the conjunction of three criteria constrains detection to 1.48\% of features.
All qualitative findings are robust to threshold choice across a 2$\times$ range (Table~\ref{tab:thresholds}), and our causal experiments (Section~\ref{sec:causal}) use an independent, percentile-based decoder-alignment method.
Prevalence is the single-token count normalized by one layer's feature width.

\begin{table}[t]
\centering
\caption{Detection thresholds and validation (GPT2-Small, 24,576 features/layer $\times$ 12 layers).}
\label{tab:thresholds}
\vspace{-2mm}
\small
\resizebox{\columnwidth}{!}{%
\begin{tabular}{@{}ccccccc@{}}
\toprule
\textbf{Gap} & \textbf{Purity} & \textbf{Word} & \textbf{Count} & \textbf{\%} & \textbf{Expl Match}\textsuperscript{$\dagger$} & \textbf{Sim Ratio} \\
\midrule
$\geq 0.2$ & $\geq 0.5$ & No & 7,171 & 29.2 & 52\% & 2.1$\times$ \\
$\geq 0.3$ & $\geq 0.6$ & No & 4,227 & 17.2 & 61\% & 3.4$\times$ \\
$\geq 0.3$ & $\geq 0.6$ & Yes & \textbf{364} & \textbf{1.48} & \textbf{64\%} & \textbf{4.7$\times$} \\
$\geq 0.4$ & $\geq 0.7$ & Yes & 148 & 0.60 & 81\% & 5.2$\times$ \\
\bottomrule
\end{tabular}}
\vspace{1mm}
{\footnotesize \textsuperscript{$\dagger$}Expl Match: fraction of features whose Neuronpedia auto-interpretability explanation contains the literal top-activating token.}
\end{table}

\subsection{Decoder-Alignment Detection}

Activation-based detection finds near-zero LlamaScope features because the TopK$\to$JumpReLU conversion alters the activation distribution.
For causal experiments requiring cross-family detection, we use \textit{decoder-alignment detection}, which replaces the activation profile with the cosine similarity between each decoder vector $\vw_i^{\text{dec}}$ and every row of the model's token embedding matrix $\mathbf{E}$.
Taking the top $k\!=\!20$ tokens by that cosine, the target token is the top-1 token, the gap ratio is computed as in Eq.~\ref{eq:gap} on the top two cosines, and a feature is selected when its gap ratio exceeds the 99th percentile of that layer's gap-ratio distribution and its top-1 cosine exceeds 0.2.
Features whose target token is whitespace or a special token are excluded, the analog of the complete-word criterion.
Purity is recorded per feature under Eq.~\ref{eq:purity} but is not a selection criterion here: the percentile gap threshold is adaptive per layer, which is what makes this detector independent of the fixed activation-based operating point.
This method requires no activation data.
The two methods are consistent where both apply: mean cosine 0.67 for single-token features and 89\% logit-lens top-token match in Section~\ref{sec:results}.
As a direct LlamaScope check, all 2{,}872 decoder-aligned single-token features across 32 layers activate on their putative token at least once in 4{,}096 evaluation sequences, with 88.3\% showing negative $\Delta$logit on ablation and 69.7\% exceeding $|\Delta\text{logit}|>0.1$.

\subsection{Geometric Analysis}

We characterize feature geometry using three metrics on decoder vectors $\{\vw_i\}_{i \in \mathcal{F}}$:

\textbf{Within-Type Similarity.}
For a feature subset $\mathcal{F}$ with decoder vectors $\{\vw_i\}_{i \in \mathcal{F}}$,
\begin{equation}
\text{sim}_{\text{within}}(\mathcal{F}) = \frac{2}{|\mathcal{F}|(|\mathcal{F}|-1)} \sum_{i < j \in \mathcal{F}} \cos(\vw_i, \vw_j).
\label{eq:withinsim}
\end{equation}
The \textit{clustering ratio} is $\text{sim}_{\text{within}}(\mathcal{F}_{\text{ST}}) / \text{sim}_{\text{within}}(\mathcal{F}_{\text{poly}})$, where $\mathcal{F}_{\text{poly}}$ is the complement of $\mathcal{F}_{\text{ST}}$ among activation-detected features at the same layer.
A ratio $>1$ indicates tighter ST clustering.

\textbf{Intrinsic Dimension.}
We want the dimension of the surface the feature vectors actually lie on, which a variance-threshold method like PCA recovers only if that surface is a linear subspace: the Levina-Bickel estimator instead reads the dimension off how quickly a point's neighbor count grows with radius---so it stays valid when the surface is curved. We use its pooled-average variant \citep{levina2004maximum}:
\begin{equation}
\hat{d} = \left[ \frac{1}{n(k_{\text{ID}}-1)} \sum_{i=1}^{n} \sum_{j=1}^{k_{\text{ID}}-1} \log \frac{r_{k_{\text{ID}}}(\vw_i)}{r_j(\vw_i)} \right]^{-1},
\label{eq:id}
\end{equation}
with $n = |\mathcal{F}|$, $k_{\text{ID}} = 10$ nearest neighbors distinct from the detection top-$k = 20$, $j$ indexing nearest-neighbor ranks $1$ through $k_{\text{ID}}-1$, and $r_j(\vw_i)$ the Euclidean distance from $\vw_i$ to its $j$-th nearest neighbor in $\mathcal{F}$.

\textbf{Cross-Layer Grassmannian Alignment.}
Let $U_\ell, U_{\ell+1} \in \R^{d \times d_{\text{PCA}}}$ have orthonormal columns spanning the top-$d_{\text{PCA}}{=}50$ principal subspaces of the decoder matrices at adjacent layers \citep{balagansky2025mechanistic, lindsey2024crosscoders}.
The alignment is
\begin{equation}
\text{align}(\ell, \ell+1) = \frac{1}{d_{\text{PCA}}} \sum_{i=1}^{d_{\text{PCA}}} \cos(\theta_i),
\label{eq:grassmann}
\end{equation}
with $\{\theta_i\}_{i=1}^{d_{\text{PCA}}}$ the principal angles from the singular value decomposition (SVD) of $U_\ell^\top U_{\ell+1}$---equivalently the singular values $\sigma_i(U_\ell^\top U_{\ell+1}) = \cos(\theta_i)$.
Values approach 1 for identical subspaces and 0 for orthogonal subspaces under a representational shift.

\section{Results}
\label{sec:results}

Table~\ref{tab:claim_evidence} maps each empirical claim to its model set, detector, layer coverage, and evidence type.

\begin{table}[h]
\centering
\caption{Claim$\to$evidence mapping.
Detector: A = activation-based; D = decoder-alignment.
Evidence: G = Geometric; Ds = Descriptive; C = Causal.}
\label{tab:claim_evidence}
\vspace{-2mm}
\scriptsize
\resizebox{\columnwidth}{!}{%
\begin{tabular}{@{}llll@{}}
\toprule
\textbf{Claim} & \textbf{Coverage} & \textbf{Det.} & \textbf{Ev.} \\
\midrule
4.7$\times$ decoder clustering & GPT2 L0 & A & G \\
Layer-0 concentration 91\% & GPT2, all 12 L & A & Ds \\
46$\times$ cross-family prevalence & 6 models, all L & A+D & Ds \\
Causal necessity (178/208 BH-sig.) & 7 cfgs, 208 L & D & C \\
Depth--necessity gradient ($\rho{=}0.97/0.70/0.81$) & G2-2B, G2-9B & D & C \\
Anchor in early layers ($\rho{=}{-}0.65$) & G2-2B & D & C \\
24\% cross-model semantic match & GPT2 $\leftrightarrow$ Gemma & A & G \\
\bottomrule
\end{tabular}}
\end{table}

\subsection{Scale-Dependent Prevalence}

Single-token feature prevalence (the fraction of a layer's features meeting all three detection criteria; Section~\ref{sec:methods}) decreases consistently with model scale within the GemmaScope/res-jb SAE family (Figure~\ref{fig:scaling_law}).
GPT2-Small (124M) yields 1.48\% single-token features (364/24{,}576, with 91\% in Layer~0), declining to 0.47\% in Gemma-2-2B (2B) and 0.14\% in Gemma-2-9B (9B).
The decline is steeper at Layer 0 (1.35\% $\rightarrow$ 0.50\% $\rightarrow$ 0.01\%), where token identity is primarily encoded.
The pattern holds only for GemmaScope/res-jb SAEs; both analyzed LlamaScope SAEs exhibit near-zero prevalence at the 8B scale. The prevalence gap co-varies with SAE family and cannot be attributed to methodology alone, since base model, tokenizer, and training data vary simultaneously.

Wider SAEs allocate more capacity to monosemantic features (prevalence increases sublinearly with width; Appendix~\ref{app:expansion}).
We report the three-point trend qualitatively; three data points cannot distinguish functional forms.
A descriptive fit with explicit ``$n\!=\!3$, descriptive only'' caveat is in Appendix~\ref{app:appendix}.
The cleanest scale comparison is within-family: Gemma-2-2B to Gemma-2-9B, with the same tokenizer and family, declines from 0.47\% to 0.14\%, while GPT2-Small extends the range without carrying inferential weight.
The Gemma-2-9B/Llama-3.1-8B contrast (Figure~\ref{fig:scaling_law}C) is a motivating cross-model observation; the within-model comparisons in Table~\ref{tab:withinmodel} are the evidentiary basis for the causal methodology claim.

\subsection{Validation}

We characterize and validate detection with three signals (Figure~\ref{fig:validation}).
\textbf{Activation profile:} Single-token features have higher max activations and peaked distributions (kurtosis 2.65 vs $-$0.08, $p<10^{-154}$), with 92\% mass on the top token vs 34\% for polysemantic: peakedness is expected given the gap-ratio criterion and confirms a clean threshold, not independent evidence.
\textbf{Geometry}, independent of activation detector: the detection criteria use only activation statistics; geometric metrics use only decoder vectors.
Single-token vectors cluster 4.7$\times$ tighter (mean cosine 0.103 in $\mathcal{F}_{\text{ST}}$ vs 0.022 in $\mathcal{F}_{\text{poly}}$) with 2$\times$ lower intrinsic dimension (60.4 vs 118.4); ratios stable across models (4.5--5.0$\times$ similarity, 1.7--2$\times$ dimension).
\textbf{Explanation}, independent of activation and geometry: \emph{explanation match} is the fraction of features whose Neuronpedia auto-interpretability explanation contains the literal case-normalized surface form of the top-activating token, 64\% for single-token vs 8\% for polysemantic \citep{ma2025revising}, widening at stricter thresholds per Table~\ref{tab:thresholds}.

We test whether single-token decoder vectors align with token embeddings, restricting to \textit{activation-detected} features to avoid circularity with decoder-alignment detection.
We define the \emph{mean cosine alignment} of a feature set $\mathcal{F}$ as the average of $\cos(\vw_i^{\text{dec}}, \mathbf{E}_{t_i^*})$ over $i \in \mathcal{F}$, where $\mathbf{E}_{t_i^*}$ is the input-embedding row for feature $i$'s top activating token $t_i^*$.
Single-token features show higher mean cosine alignment than polysemantic (0.67 vs 0.39, $p<10^{-42}$), decreasing with layer depth as expected for increasingly abstract representations.
With the logit lens \citep{nostalgebraist2020logitlens, bloom2024logitlens}, the \emph{logit-lens match} indicator is 1 if $\arg\max_t (\vw_i^{\text{dec}} \cdot \mathbf{U}_t) = t_i^*$, 0 otherwise; 89\% match for single-token vs 12\% for polysemantic:
this bidirectional alignment, from embedding space and toward unembedding, validates that single-token features function as lexical identity detectors.

\subsection{Layer-0 Concentration and Representational Shift}

\begin{figure*}[t]
    \centering
    \includegraphics[width=\textwidth]{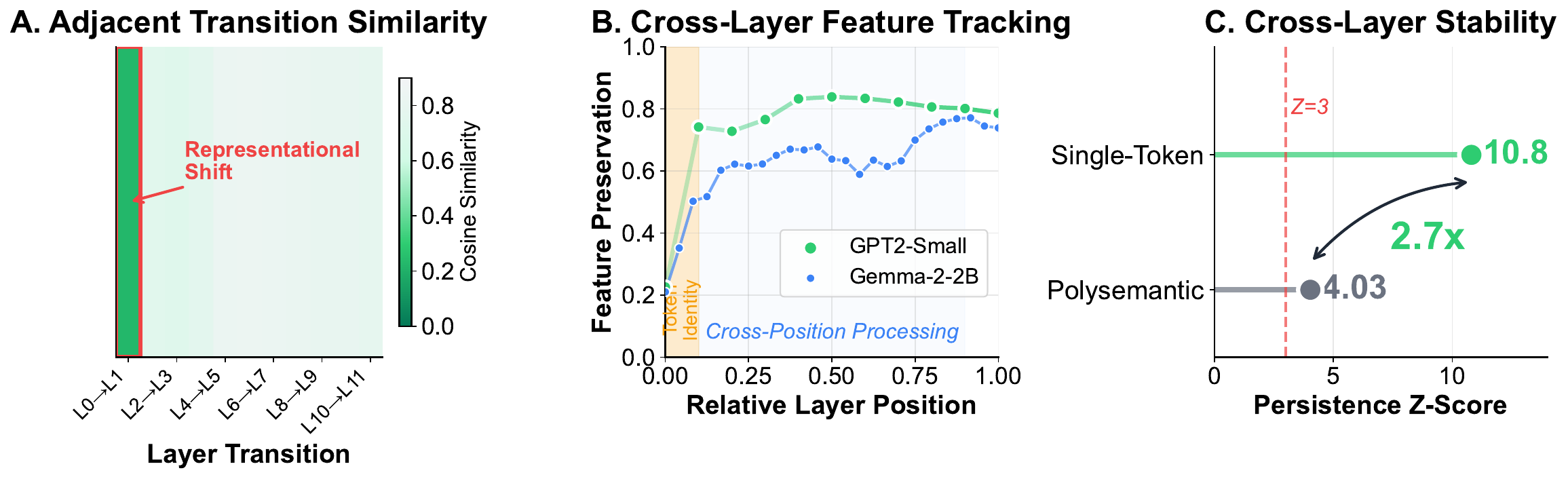}
    \caption{\textbf{Cross-layer dynamics.} (A) Cross-layer similarity matrix (GPT2): L0$\to$L1 shows sharp transition (highlighted), then stabilizes. (B) Feature preservation across layers: both models show low L0$\to$L1 similarity, then recover. (C) Persistence Z-scores: single-token features show 2.7$\times$ higher cross-layer persistence (10.7 vs 4.0).}
    \label{fig:layer}
\end{figure*}

In GPT2-Small, the majority of single-token features concentrate in Layer 0 (Figure~\ref{fig:layer}; per-layer breakdown in Table~\ref{tab:layer_dist}, Appendix).
This concentration precedes a sharp representational shift: cross-layer feature tracking shows 86.8\% of L0 features have no close L1 match at the 0.3 cosine threshold, 3$\times$ the mean random-pair similarity, while L1$\rightarrow$L2 shows only 16.0\% transformation and later transitions maintain 60--70\% feature persistence.
Gemma-2-2B confirms this pattern: L0$\rightarrow$L1 transformation is 82.8\%, then L1$\rightarrow$L2 drops to 51.9\%, with later layers showing $<$5\% transformation.
This aligns with Layer 0 encoding token identity before attention enables cross-position flow.
Grassmannian alignment (Table~\ref{tab:grassmann}) quantifies the shift: GPT2 L0$\to$L1 alignment is 0.26 while later transitions stabilize at $>$0.9.
Gemma L0$\to$L1 alignment is similarly low at 0.23 under its distributed single-token pattern.
Intrinsic dimension analysis (full table in Appendix~\ref{app:layer_geometry}): single-token features in GPT2 L0 occupy a lower-dimensional manifold than polysemantic features.
Despite the L0$\to$L1 shift, single-token features show 2.7$\times$ higher cross-layer persistence than polysemantic features at Z-score 10.7 vs 4.0, $p < 10^{-73}$.

\begin{table}[t]
\centering
\caption{Grassmannian alignment between adjacent layers.}
\label{tab:grassmann}
\vspace{-2mm}
\small
\resizebox{\columnwidth}{!}{%
\begin{tabular}{@{}l|cccccc@{}}
\toprule
\textbf{Model} & L0$\to$1 & L1$\to$2 & L2$\to$3 & L3$\to$4 & L4$\to$5 & Mean(L5+) \\
\midrule
GPT2-Small & 0.26 & 0.95 & 0.96 & 0.95 & 0.95 & 0.92 \\
Gemma-2-2B & 0.23 & 0.36 & 0.42 & 0.34 & 0.40 & 0.51 \\
\bottomrule
\end{tabular}}
\end{table}

High-persistence pairs at $Z>50$ preserve semantic meaning, 0.68 vs 0.16 in explanation similarity, replicating in Gemma-2-2B at 0.61 vs 0.32.
Single-token features also show 1.42$\times$ smaller direction change at L1$\rightarrow$L2 and near-zero co-occurrence at Jaccard $<$0.001, partitioning vocabulary into non-overlapping regions.

\subsection{Token Characterization}

Single-token features are enriched 2.5$\times$ for mid-frequency tokens at rank 1k--10k, 66.5\% vs 26.7\%.
Content nouns at 52\%, proper nouns at 8.4\%, and function words at 4.6\% dominate (Table~\ref{tab:semantic}).
Ablation damage varies across categories under Kruskal-Wallis $p = 4 \times 10^{-72}$.
The ``Goldilocks zone'' pattern is tested against three chi-squared nulls with $df = 5$: uniform, vocab-frequency-proportional, and polysemantic-matched.
All three are rejected ($\chi^2 \geq 206.4$, all $p < 10^{-41}$), ruling out simple allocation models.

Despite different tokenizers, 24\% of GPT2 single-token features have a Gemma counterpart with sentence-embedding cosine $>$0.7 over auto-interpretability explanations (Table~\ref{tab:correspondence}): partial semantic convergence across models, measured at the explanation-text level rather than the geometric feature level.

\begin{table}[t]
\centering
\caption{Semantic category distribution. $\Delta\text{logit}$: mean logit change on ablation (more negative = more damage). Damaged: fraction with $|\Delta\text{logit}| > 0.1$.}
\label{tab:semantic}
\vspace{-2mm}
\small
\resizebox{\columnwidth}{!}{%
\begin{tabular}{@{}l|cc|cc|cc@{}}
\toprule
& \multicolumn{2}{c|}{\textbf{Prevalence}} & \multicolumn{2}{c|}{\textbf{Causal ($N\!=\!26{,}594$)}} & \multicolumn{2}{c}{\textbf{Freq Rank}} \\
\textbf{Category} & Count & \% & $\Delta\text{logit}$ & Damaged & GPT2 & Gemma \\
\midrule
Numbers/Digits & 176 & 0.7\% & $\mathbf{-0.522}$ & 58\% & 0.5k--5k & 1k--8k \\
Function Words & 1,225 & 4.6\% & $-0.220$ & 38\% & 50--500 & 80--800 \\
Proper Nouns & 2,222 & 8.4\% & $-0.070$ & 13\% & 1.2k--4.5k & 2.1k--8.2k \\
Content Nouns & 13,827 & 52.0\% & $-0.063$ & 13\% & 1.5k--6.1k & 2.8k--9.1k \\
Abbreviations & 958 & 3.6\% & $-0.033$ & 9\% & varies & varies \\
\bottomrule
\end{tabular}}
\\[2pt]
\footnotesize{Kruskal-Wallis $H = 496.5$, $p = 4 \times 10^{-72}$.
$N = 26{,}594$ across 6 models.
Top 5 categories shown (69\% of features); remaining 31\% span minor categories.}
\end{table}

\subsection{Causal Validation}
\label{sec:causal}

To establish functional necessity beyond correlational evidence, we perform zero-ablation interventions across eight model$\times$SAE combinations (Table~\ref{tab:causal}).
For a feature $i$ active with value $f_i > 0$ at a (sequence, position) pair, we replace the residual-stream activation $\mathbf{a}$ with $\mathbf{a} - f_i \cdot \mathbf{w}_i^{\text{dec}}$, run the modified forward pass, and record \emph{ablation damage} $\Delta\text{logit}_i = z_{\text{ablated}}(t_i^*) - z_{\text{clean}}(t_i^*)$, where $z(t)$ is the pre-softmax logit the model assigns to the top activating token $t_i^*$, averaged over all positions where $f_i > 0$---we exclude features with fewer than five active positions.
We test layer-level significance with a one-sided Mann-Whitney $U$ on signed $\Delta\text{logit}$ (alternative: single-token ablation shifts the target-token logit more negatively than matched controls), BH-corrected at $p<0.05$ globally across all 208 layer tests.
Features are identified via decoder-alignment detection, Section~\ref{sec:methods}.
Detection consistency with the activation-based detector is supported by 89\% logit-lens match and 0.67 mean cosine.
We take activation positions from 4{,}096 sequences of 128 tokens of OpenWebText \citep{gokaslan2019openwebtext}---each model tokenizes its own copy of the raw text.
Each single-token feature gets a \emph{size-matched random control} drawn uniformly from non-single-token features of the same SAE---restricted to controls whose mean active activation falls within a 2$\times$ range of the ST feature's.

\textbf{Necessity.} Single-token feature ablation reduces target token logits at BH-significant levels across all eight conditions (Table~\ref{tab:causal}).
Within the four full-layer experiments on Gemma models, 77/102 tested layers yield significant logit reduction under Mann-Whitney $U$ globally BH-corrected at $p < 0.05$: 50/51 Gemma-2-2B layers and 27/51 Gemma-3-1B layers, concentrated in later layers.
Three additional full-depth configurations extend this: Llama-3.1-8B $\times$ LlamaScope at 31/32 significant layers and prevalence 0.27\%, Gemma-2-9B $\times$ GemmaScope at 42/42, and DeepSeek-R1 $\times$ LlamaScope at 28/32.
Per-condition peak $p$ values and feature counts are in Table~\ref{tab:causal}.
Total full-layer causal coverage is 208 layers across 7 model$\times$SAE configurations.
This effect is not explained by activation magnitude: even in the lowest activation quartile, single-token features cause more damage than magnitude-matched random controls ($p < 0.0001$, rank-biserial $r = 0.27$--$0.43$; Appendix~\ref{app:magnitude}).
Nor is it an artifact of the decoder-alignment selection: against an embedding-alignment-matched null population, single-token necessity remains significant at eight of nine layer-configurations (Appendix~\ref{app:geonull}).

\begin{figure*}[t]
    \centering
    \includegraphics[width=\textwidth]{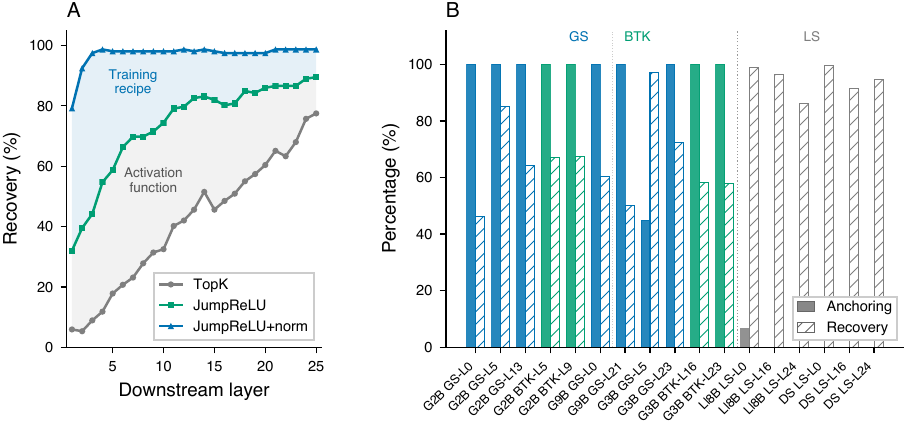}
    \caption{\textbf{Causal structure across SAE families.}
    (A) Recovery curves for three SAE types on Gemma-2-2B L1: activation function and training recipe each contribute to the recovery gap.
    (B) Anchoring (solid) and recovery (hatched) across 17 sampled model$\times$layer conditions; per-layer breakdown for the full 208-layer coverage is in Table~\ref{tab:causal}.
    GemmaScope (blue) and BatchTopK (green) show high anchoring with moderate recovery; LlamaScope (gray) shows near-zero anchoring with high recovery.}
    \label{fig:causal}
\end{figure*}

\textbf{Anchoring and redundancy.} A feature at source layer $\ell$ \emph{anchors} downstream layer $\ell' > \ell$ if zero-ablation at $\ell$ produces a BH-significant change in the logit-lens top-token readout at $\ell'$, using one-sided Mann-Whitney $U$ against magnitude-matched controls at $p < 0.05$.
A source layer \emph{anchors $\geq 1$ downstream layer} if at least one $\ell' > \ell$ meets this criterion under global BH correction.
Anchoring is consistent for GemmaScope and BatchTopK at 92--100\% of source layers and sparser for LlamaScope, where Llama-3.1-8B anchors 31\% and DeepSeek-R1 34\% of source layers at full 32-layer depth; the residual anchoring on Llama-3.1-8B concentrates in early layers L0--L9.
Same-layer \emph{recovery} is the fraction of features whose mean rank for $t_i^*$ under the same-layer logit lens stays within twice its pre-ablation value (floor 5) \emph{after} ablation.
High recovery indicates \emph{local redundancy} where other features compensate; low recovery indicates critical reliance on the ablated feature.
Recovery mirrors anchoring: GemmaScope 62--71\%, LlamaScope 96--98\%, with Gemma-3-1B GemmaScope at 91\% is the exception (Figure~\ref{fig:causal}).

This SAE family split in causal structure parallels the prevalence split: SAE training methodology shapes not only which features are detected but their degree of causal importance.

\textbf{Layer depth dissociation.} Necessity and anchoring show opposite layer profiles: necessity damage ($|\Delta\text{logit}|$) rises monotonically with depth (Spearman $\rho = 0.97$ for BatchTopK, $0.70$ for GemmaScope on Gemma-2-2B; $p < 0.001$), with late layers showing 13--30$\times$ more damage than early layers.
In contrast, anchor damage (downstream propagation) is concentrated in early layers: early-layer ablations cause 4--16$\times$ more total downstream disruption than late-layer ablations (Table~\ref{tab:layer_dissociation}).
This dissociation shows complementary roles: early features are propagation anchors whose ablation cascades through the network, while late features directly shape the output distribution.
BatchTopK shows a near-perfect monotonic necessity gradient ($\rho = 0.97$) while GemmaScope has a noisier profile ($\rho = 0.70$): SAE families distribute causal load across layers.

\textbf{Cross-architecture validation.} Gemma-3-1B reproduces the pattern: anchoring holds across 92--96\% of layers, necessity is BH-significant in 27/51 layers concentrated late, and BatchTopK on the same model recovers 18/25 of them (Table~\ref{tab:causal}).
The weaker Gemma-3-1B GemmaScope replication (9/26 vs 26/26 on Gemma-2-2B) is consistent with smaller effect sizes rather than a methodology breakdown.
Per-layer mean $|\Delta\text{logit}|$ on Gemma-3-1B GemmaScope is order-of-magnitude smaller than Gemma-2-2B GS at matched $n_{\text{ST}}$ (Table~\ref{tab:layer_dissociation}).
On Gemma-3-1B BatchTopK, where effect sizes recover, 18/25 layers are BH-significant.

\begin{table}[t]
\centering
\caption{Causal validation across 8 model$\times$SAE conditions; the seven full-depth configurations contribute the 208 layers reported in the main text (GPT2-Small is a single-layer condition). Nec.\ sig: layers significant under a single global BH correction across all 208 tests. Peak $p$: strongest per-layer one-sided Mann-Whitney result on the signed statistic. Recovery: fraction of features whose target-token rank after ablation stays within twice its pre-ablation rank (floor 5); blue cells mark the anchored regime, rust cells the locally redundant LlamaScope regime.}
\label{tab:causal}
\vspace{-2mm}
\small
\resizebox{\columnwidth}{!}{%
\begin{tabular}{@{}llccccccc@{}}
\toprule
\textbf{Model} & \textbf{SAE Family} & \textbf{Layers} & $n_\text{ST}$ & \textbf{Nec.\ sig} & \textbf{Peak $p$} & \textbf{Anchor ($\geq$1)} & \textbf{Avg anc.} & \textbf{Recovery} \\
\midrule
GPT2-Small & res-jb (ReLU) & 1/12 & 220 & 1/1 & $2.5\mathrm{e}{-80}$ & 1/1 & 100\% & \hlanchored{26.8\%} \\
\midrule
Gemma-2-2B & GemmaScope (JR) & 26/26 & 51--157 & \textbf{26/26} & $1.4\mathrm{e}{-34}$ & 25/26 & 50\% & \hlanchored{70.6\%} \\
Gemma-2-2B & BatchTopK & 25/25 & 64--322 & \textbf{24/25} & $\mathbf{2.9\mathrm{e}{-67}}$ & 25/25 & 52\% & \hlanchored{68.0\%} \\
Gemma-2-9B & GemmaScope (JR) & \textbf{42/42} & 42--157 & \textbf{42/42} & $\mathbf{6.1\mathrm{e}{-26}}$ & 41/42 & 50\% & \hlanchored{62.1\%} \\
\midrule
Gemma-3-1B & GemmaScope (JR) & 26/26 & 56--150 & 9/26 & $7.8\mathrm{e}{-8}$ & 24/26 & 25\% & 90.8\% \\
Gemma-3-1B & BatchTopK & 25/25 & 25--236 & \textbf{18/25} & $1.0\mathrm{e}{-29}$ & 24/25 & 46\% & \hlanchored{68.4\%} \\
\midrule
Llama-3.1-8B & LlamaScope (TopK$\to$JR) & \textbf{32/32} & 1--328 & \textbf{31/32} & $\mathbf{2.3\mathrm{e}{-141}}$ & 10/32 & 23\% & \hlredundant{97.7\%} \\
DeepSeek-R1 & LlamaScope (TopK$\to$JR) & \textbf{32/32} & 3--321 & \textbf{28/32} & $\mathbf{2.0\mathrm{e}{-13}}$ & 11/32 & 14\% & \hlredundant{95.5\%} \\
\bottomrule
\end{tabular}}
\end{table}

\begin{table}[t]
\centering
\caption{Layer depth dissociation: necessity increases with depth while anchor damage concentrates in early layers. Q1/Q4: first/last quartile of layers.}
\label{tab:layer_dissociation}
\vspace{-2mm}
\small
\resizebox{\columnwidth}{!}{%
\begin{tabular}{@{}ll|cc|cc@{}}
\toprule
& & \multicolumn{2}{c|}{\textbf{Necessity ($\Delta$logit)}} & \multicolumn{2}{c}{\textbf{Anchor (downstream)}} \\
\textbf{Model} & \textbf{SAE} & Q1 (early) & Q4 (late) & Q1 (early) & Q4 (late) \\
\midrule
Gemma-2-2B & BatchTopK & $-0.006$ & \hlbest{$-0.182$} & \hlbest{64k} & 4k \\
Gemma-2-2B & GemmaScope & $-0.020$ & \hlbest{$-0.262$} & \hlbest{29k} & 4k \\
Gemma-3-1B & BatchTopK & $-0.070$ & \hlbest{$-0.229$} & \hlbest{7k} & 1k \\
Gemma-3-1B & GemmaScope & $-0.010$ & \hlbest{$-0.043$} & \hlbest{0.4k} & 0.1k \\
\midrule
\multicolumn{2}{@{}l|}{\textbf{Spearman $\rho$ (depth$\leftrightarrow$nec.)}} & \multicolumn{2}{c|}{0.61--0.97 (Gemma-3-1B GS to Gemma-2-2B BTK; $p < 0.001$)} & \multicolumn{2}{c}{$-0.65$ ($p < 0.001$)} \\
\bottomrule
\end{tabular}}
\end{table}

Table~\ref{tab:withinmodel} (full breakdowns in Appendix~\ref{app:paired}, \ref{app:controlled}) jointly indicates that the activation function alone does not explain the cross-family pattern: token-matched ($N\!=\!627$) shows BatchTopK $>$ GemmaScope ($p = 1.2 \times 10^{-18}$, $r\!=\!0.36$), but the activation-function-isolated controlled comparison on the same model/layer/width ($N\!=\!142$) shows the opposite direction (JumpReLU $>$ TopK, $p\!=\!0.036$); opposite signs from the same model leave us with training-recipe factors as residual candidates (discussed in Section~\ref{sec:discussion}).

\begin{table}[t]
\centering
\caption{Within-model decompositions of the cross-family causal gap.
Top: token-matched paired (BTK vs GS, $N\!=\!627$).
Bottom: controlled activation-function on Gemma-2-2B L1 ($N\!=\!142$, width 18k).}
\label{tab:withinmodel}
\vspace{-2mm}
\small
\resizebox{\columnwidth}{!}{%
\begin{tabular}{@{}lcccc@{}}
\toprule
\textbf{Comparison} & \textbf{Metric} & \textbf{$p$} & \textbf{Effect size} & \textbf{Direction} \\
\midrule
\multicolumn{5}{@{}l}{\textit{Token-matched paired ($N\!=\!627$, Gemma-2-2B + Gemma-3-1B):}} \\
BatchTopK vs GemmaScope & Anchor ($\Delta$logit-lens) & $1.2 \times 10^{-18}$ & $r\!=\!0.36$ & \textbf{BTK $>$ GS} \\
BatchTopK vs GemmaScope & Necessity ($\Delta$rank) & $0.004$ & $r\!=\!0.12$ & \textbf{BTK $>$ GS} \\
\midrule
\multicolumn{5}{@{}l}{\textit{Controlled activation function ($N\!=\!142$, Gemma-2-2B L1, width 18k):}} \\
TopK vs JumpReLU & Necessity (mean $\Delta$rank) & $0.036$ & $r\!=\!0.20$ & \textbf{JR $>$ TopK} \\
TopK vs JumpReLU & Necessity (max $\Delta$rank) & $0.003$ & $r\!=\!0.27$ & \textbf{JR $>$ TopK} \\
\bottomrule
\end{tabular}}
\end{table}

\subsection{Factors and Robustness}

Extending to all six models (Figure~\ref{fig:scaling_law}B), GemmaScope/res-jb prevalence is 0.01--1.35\% in Layer 0 while LlamaScope is near-zero at $<$0.01\%.
The 46$\times$ family contrast exceeds the within-family scale trend, with GemmaScope gap ratios up to 0.76 vs LlamaScope's 0.17.
An eight-architecture comparison on Gemma-2-2B L12 (Appendix~\ref{app:expansion}; Table~\ref{tab:multiarch}) shows prevalence insensitive to activation function at mid-layers (0.92--0.99\%): single-token features are primarily pre-compositional.
Threshold ablations (Table~\ref{tab:thresholds}) confirm the geometric signatures are stable across operating points.
Our causal experiments (Section~\ref{sec:causal}) use independent percentile-based decoder-alignment detection.

\section{Discussion}
\label{sec:discussion}

\paragraph{Geometric Findings.}
The 4.7$\times$ tighter decoder clustering, 2$\times$ lower intrinsic dimension, and 1.72$\times$ higher embedding alignment ($p < 10^{-42}$) provide activation-independent validation.
All three support the linear representation hypothesis \citep{park2023linear, park2024geometry, li2024geometry, engels2024not}.
Similarity ratios are stable across models (4.5--5.0$\times$ at L0; mechanism figure in Appendix~\ref{app:figures}).
And cross-model analysis reveals zero surface-form overlap but 24\% semantic correspondence (Appendix Table~\ref{tab:correspondence}).
That localizes the tokenizer-invariant boundary at the single-token endpoint, and it complements \citet{lan2024universal}'s universal feature-space findings \citep{templeton2024scaling, gao2025topksae, venhoff2024sage}.

\paragraph{Layerwise Dynamics.}
GPT2's Layer-0 concentration (91\%) preceding low L0$\to$L1 alignment (0.26) supports Layer 0 encoding token identity before attention enables cross-position flow \citep{elhage2021mathematical, ameisen2025circuit}.
The L0 concentration is model-specific, though.
Gemma models show distributed L0--L4 patterns.
The depth dissociation (Section~\ref{sec:causal}) shows complementary roles: necessity damage scales with depth ($\rho = 0.97$ BatchTopK, $0.70$ GemmaScope at 2B; $0.81$ at 9B) while anchoring concentrates in early layers ($\rho = -0.65$).
This refines prior cross-layer SAE tracking work \citep{balcells2024evolution, balagansky2025mechanistic}, which characterize persistence or subspace motifs but do not separate necessity from anchor profiles.
Gemma-3-1B GemmaScope (91\% recovery) is an exception.
But BatchTopK on the same model recovers at 68.4\%, so the SAE-family ordering still holds at 1B.

\paragraph{SAE Family Comparison.}
The 46$\times$ cross-family prevalence contrast exceeds the roughly 10$\times$ within-family scale trend, and the causal contrast follows the same family split.
But the cross-family comparison varies training data, dictionary width, training recipe, and post-hoc conversion.
So we treat within-model evidence as load-bearing and the 46$\times$ number as a magnitude bound, consistent with \citet{paulo2025sparseautoencoders, leask2025canonical, chanin2024absorption, locatello2019challenging}.
Across 208 full-layer tests, the anchoring split persists: GemmaScope/BatchTopK anchor 92--100\% of source layers while LlamaScope anchors 31--34\% across Llama-3.1-8B and DeepSeek-R1.
And the effect is category-dependent at the token level, with domain-specific tokens converging at 93\% and function words at 29\% (Appendix~\ref{app:cross_sae}).
The within-model picture dissociates per Table~\ref{tab:withinmodel}: token-matched BatchTopK $>$ GemmaScope at $N\!=\!627$, $p = 1.2 \times 10^{-18}$, but the activation-function-isolated $N\!=\!142$ comparison shows JumpReLU $>$ TopK at $p\!=\!0.036$.
Opposite signs from the same model leave training corpus, dictionary width, and recipe as residual candidates \citep{geiger2025causal, hindupur2025projecting, braun2024e2e, makelov2024principled}.

\section{Conclusion}

At the single-token endpoint, where ground truth is unambiguous, the causal role of a feature depends on which SAE produced it.
We ablate a single-token feature and the target token's logit falls.
That holds across six transformer language models and three SAE families, and depth controls whether the damage cascades downstream or shapes the output directly.
And the same token can be causally anchored under one SAE family yet locally redundant under another.
So cross-SAE interpretability claims must control for training methodology, not just activation function or scale.

For practitioners this implies three concrete steps.
First, re-verify causal claims when switching SAE families.
A feature's necessity under one family cannot be assumed to transfer, even on the same base model, so steering and editing pipelines should re-run ablation checks under the family they deploy.
Second, profile a checkpoint's causal behavior directly rather than inferring it from family name or activation function, because anchoring and recovery statistics are a safer guide than either label.
Third, add per-feature causal necessity to SAE evaluation.
Two SAEs trained on the same base model can differ in causal structure, and current benchmarks such as SAEBench \citep{karvonen2025saebench} do not directly measure that dimension.

\section*{Limitations}

Our cross-family comparisons co-vary training data, dictionary width, recipe, and post-hoc conversion.
The within-model pairings on Gemma-2-2B and Gemma-3-1B and the activation-function-controlled comparison partially constrain the recipe factor.
All four full-depth configurations show consistent results.
But Gemma-3-1B at 26 layers shows weaker effects.
Activation-based detection uses a fixed operating point.
And conclusions are robust across a 2$\times$ threshold range (Table~\ref{tab:thresholds}), while the causal experiments use independent percentile-based decoder-alignment detection.
The single-token endpoint is the diagnostic case.
We leave multi-token spans and compositional features to future work, via span-embedding alignment.

\bibliography{references}

\appendix
\onecolumn
\section{Appendix}
\label{app:appendix}

\subsection{Model and SAE Configurations}

\begin{table}[h]
\centering
\caption{Models and SAE configurations analyzed.
SAE Family reflects training methodology.
Community SAEs (\dag) used in controlled comparison (Section~\ref{sec:results}).}
\label{tab:models_appendix}
\vspace{-2mm}
\small
\begin{tabular}{@{}lccccccc@{}}
\toprule
\textbf{Model} & \textbf{Params} & \textbf{$d_{\text{model}}$} & \textbf{Layers} & \textbf{Feat/L} & \textbf{SAE Type} & \textbf{Family} & \textbf{Total} \\
\midrule
GPT2-Small & 124M & 768 & 12 & 24,576 & ReLU & res-jb & 294,912 \\
Gemma-2-2B & 2B & 2,304 & 26 & 16,384 & JumpReLU & GemmaScope & 425,984 \\
Gemma-2-9B & 9B & 3,584 & 42 & 16,384 & JumpReLU & GemmaScope & 688,128 \\
Gemma-3-1B & 1B & 1,152 & 26 & 16,384 & JumpReLU & GemmaScope & 425,984 \\
Llama-3.1-8B & 8B & 4,096 & 32 & 32,768 & TopK & LlamaScope & 1,048,576 \\
DeepSeek-R1 & 8B & 4,096 & 32 & 32,768 & TopK & LlamaScope & 1,048,576 \\
\midrule
\multicolumn{8}{@{}l}{\textit{BatchTopK SAEs for budget pressure comparison:}} \\
Chanind & 2B & 2,304 & 25 & 32,768 & BatchTopK & Community & 819,200 \\
Chanind & 1B & 1,152 & 25 & 32,768 & BatchTopK & Community & 819,200 \\
\midrule
\multicolumn{8}{@{}l}{\textit{Community SAEs for controlled comparison (Gemma-2-2B L1):}} \\
Chanind\dag & 2B & 2,304 & L1 & 18,432 & TopK ($k\!=\!25$) & Community & 18,432 \\
Chanind\dag & 2B & 2,304 & L1 & 18,432 & JumpReLU & Community & 18,432 \\
GemmaScope & 2B & 2,304 & L1 & 16,384 & JumpReLU & GemmaScope & 16,384 \\
\midrule
\multicolumn{8}{@{}l}{\textit{SAEBench architectures (Gemma-2-2B L12, Section~\ref{sec:results}):}} \\
SAEBench & 2B & 2,304 & L12 & 16,384 & 8 variants & SAEBench & 131,072 \\
\bottomrule
\end{tabular}
\end{table}

\subsection{Expansion Factor Analysis}
\label{app:expansion}

SAE expansion factor affects prevalence: with GPT2-Small SAEs with widths 4$\times$, 8$\times$, 16$\times$, and 32$\times$ the model dimension, single-token prevalence increases sublinearly: from 0.8\% at 4$\times$ to 1.48\% at 32$\times$, following approximately $\text{ST\%} \propto W^{0.3}$.
This suggests wider SAEs allocate more capacity to monosemantic features, but the rate of increase diminishes.
Both geometric signatures---the similarity ratio and the dimension ratio---remain stable across widths; single-token features maintain their distinct structure regardless of SAE capacity.

\subsection{Scaling Trend Fit Details}

A descriptive power-law fit $\text{ST\%} \propto N^{\alpha}$ across three GemmaScope/res-jb models ($n\!=\!3$) yields $\alpha = -0.51 \pm 0.08$ for all-layer analysis ($R^2 = 0.97$) and $\alpha = -1.33$ for Layer 0 only ($R^2 = 0.73$): we emphasize that three data points cannot distinguish among functional forms---this fit is primarily descriptive: the steeper Layer 0 exponent reflects concentrated single-token encoding in early layers.
The geometric similarity ratio (4.5--5.0$\times$ at L0) remains stable across model families.
Intrinsic dimension ratios increase with depth (1.7--2$\times$ at L0, $\sim$14$\times$ at mid layers; Table~\ref{tab:id}), reflecting greater dimensionality divergence---representations become more compositional with depth.

We fit the power law $\text{ST\%} = a \cdot N^{\alpha}$ using weighted least squares on log-transformed data.
Bootstrap resampling (1000 iterations) yields confidence intervals for each feature type.

\begin{table}[h]
\centering
\caption{Scaling trend parameters for different feature types ($n\!=\!3$; primarily descriptive).}
\label{tab:scaling_params}
\vspace{-2mm}
\small
\begin{tabular}{@{}lcccc@{}}
\toprule
\textbf{Feature Type} & $\alpha$ & \textbf{Boot.\ range ($n\!=\!3$)} & $R^2$ & \textbf{Intercept} \\
\midrule
Single-Token & $-0.51$ & $[-0.59, -0.43]$ & 0.97 & 4.82 \\
Morpheme & $-0.42$ & $[-0.51, -0.33]$ & 0.94 & 5.14 \\
Concept & $-0.31$ & $[-0.42, -0.20]$ & 0.89 & 4.21 \\
Polysemantic & $+0.08$ & $[-0.02, +0.18]$ & 0.42 & 91.3 \\
\bottomrule
\end{tabular}
\end{table}

\textbf{Interpretation:} The exponent $\alpha$ indicates how feature prevalence scales with model size.
Negative $\alpha$ means the feature type becomes rarer in larger models.
Single-token features show the steepest decline ($\alpha = -0.51$), meaning each 10$\times$ increase in parameters reduces their prevalence to roughly one-third ($10^{-0.51} \approx 0.31$).
Morpheme and concept features decline more slowly, while polysemantic features remain constant ($\alpha \approx 0$), absorbing capacity freed by diminishing monosemantic features.
The intercept represents log-prevalence at $N=1$, used only for curve fitting.

\subsection{Layer Distribution and Alignment Notes}
\label{app:layer_geometry}

\begin{table}[h]
\centering
\caption{Intrinsic dimension by layer and feature type.}
\label{tab:id}
\vspace{-2mm}
\scriptsize
\begin{tabular}{@{}l|cc|cc|c@{}}
\toprule
& \multicolumn{2}{c|}{\textbf{GPT2-Small}} & \multicolumn{2}{c|}{\textbf{Gemma-2-2B}} & \\
\textbf{Layer} & ST & Poly & ST & Poly & \textbf{Ratio} \\
\midrule
L0 & 60.4 & 118.4 & 106.5 & 180.4 & 1.7--2$\times$ \\
Mid & 28.1 & 398.2 & 35.8 & 498.7 & $\sim$14$\times$ \\
Final & 31.4 & 421.5 & 42.1 & 512.3 & 12--13$\times$ \\
\bottomrule
\end{tabular}
\end{table}

\begin{table}[h]
\centering
\caption{Representative single-token features (GPT2-Small Layer 0).}
\label{tab:examples}
\vspace{-2mm}
\small
\begin{tabular}{@{}lcccl@{}}
\toprule
\textbf{Token} & \textbf{Gap} & \textbf{Purity} & \textbf{Sparsity} & \textbf{Top Activations} \\
\midrule
commenting & 0.42 & 0.81 & 0.018\% & commented, comment, comments \\
Parks & 0.38 & 0.64 & 0.024\% & parks, Park, parking, Recreation \\
sovereign & 0.45 & 0.82 & 0.012\% & Sovereign, sovereignty, sovereigns \\
follow & 0.40 & 1.00 & 0.021\% & Follow, follows, followed, following \\
triangle & 0.51 & 0.91 & 0.009\% & triangles, Triangle, triangular \\
Manhattan & 0.48 & 0.87 & 0.011\% & manhattan, MANHATTAN \\
climbing & 0.39 & 0.73 & 0.019\% & climb, climbed, climbs, climber \\
Egyptian & 0.44 & 0.79 & 0.015\% & Egypt, Egyptians, egyptian \\
\bottomrule
\end{tabular}
\end{table}

Tables~\ref{tab:grassmann} and \ref{tab:id} in the main text show Grassmannian alignment and intrinsic dimension.
Table~\ref{tab:layer_dist} shows per-layer single-token feature counts.
GPT2-Small concentrates 91\% of single-token features (331/364) in Layer~0, with exponential decay.
Gemma-2-2B shows a distributed pattern under GemmaScope (no layer exceeds 6\%), while BatchTopK concentrates features in later layers.
Gemma-3-1B shows intermediate behavior.
Feature counts reflect decoder-alignment detection for Gemma models and activation-based detection for GPT2.

\begin{table}[h]
\centering
\caption{Single-token feature count by layer (L0--L11).}
\label{tab:layer_dist}
\vspace{-2mm}
\scriptsize
\begin{tabular}{@{}r|rr|rr|rr|rr@{}}
\toprule
& \multicolumn{2}{c|}{\textbf{GPT2}} & \multicolumn{2}{c|}{\textbf{G2-2B GS}} & \multicolumn{2}{c|}{\textbf{G2-2B BTK}} & \multicolumn{2}{c}{\textbf{G3-1B GS}} \\
\textbf{L} & $n$ & \% & $n$ & \% & $n$ & \% & $n$ & \% \\
\midrule
0 & 331 & 1.35 & 82 & 0.50 & 260 & 0.79 & 89 & 0.54 \\
1 & 17 & 0.07 & 156 & 0.95 & 64 & 0.20 & 59 & 0.36 \\
2 & 4 & 0.02 & 157 & 0.96 & 164 & 0.50 & 56 & 0.34 \\
3 & 3 & 0.01 & 149 & 0.91 & 188 & 0.57 & 64 & 0.39 \\
4 & 2 & 0.01 & 141 & 0.86 & 211 & 0.64 & 109 & 0.67 \\
5 & 2 & 0.01 & 134 & 0.82 & 298 & 0.91 & 142 & 0.87 \\
6 & 2 & 0.01 & 131 & 0.80 & 322 & 0.98 & 141 & 0.86 \\
7 & 1 & 0.00 & 121 & 0.74 & 316 & 0.96 & 138 & 0.84 \\
8 & 1 & 0.00 & 118 & 0.72 & 317 & 0.97 & 150 & 0.92 \\
9 & 0 & 0.00 & 99 & 0.60 & 318 & 0.97 & 141 & 0.86 \\
10 & 1 & 0.00 & 103 & 0.63 & 242 & 0.74 & 148 & 0.90 \\
11 & 0 & 0.00 & 110 & 0.67 & 225 & 0.69 & 144 & 0.88 \\
\midrule
\multicolumn{2}{@{}l|}{\textbf{Total}} & 364 & & 2,626 & & 4,574 & & 3,206 \\
\multicolumn{2}{@{}l|}{\textbf{L0\%}} & 91\% & & 3\% & & 6\% & & 3\% \\
\bottomrule
\end{tabular}
\end{table}

Grassmannian alignment is computed between top-$d_{\text{PCA}}{=}50$ principal subspaces of decoder matrices at adjacent layers using $\text{align}(U, V) = \frac{1}{d_{\text{PCA}}}\sum_{i=1}^{d_{\text{PCA}}} \sigma_i(U^\top V)$, where the singular values $\sigma_i(U^\top V) = \cos(\theta_i)$ recover the principal-angle form of eq.~\ref{eq:grassmann}.
Values near 0 indicate orthogonal subspaces, i.e.\ major representational shift; values near 1 indicate aligned subspaces with gradual change.
GPT2's low L0$\to$L1 alignment (0.26) coincides with single-token feature concentration (91\% in L0).
After this shift, alignment stabilizes at $>$0.9.
Gemma shows similarly low L0$\to$L1 alignment (0.23).

Single-token features occupy manifolds with intrinsic dimension 60--107, while polysemantic features span 118--180 dimensions, a 1.7--2$\times$ ratio.
MLE dimension estimation uses $k_{\text{ID}}\!=\!10$ nearest neighbors (Eq.~\ref{eq:id}) following \citet{levina2004maximum}.

\subsection{Multi-Architecture Comparison}

\begin{wraptable}{r}{0.45\textwidth}
\vspace{-4mm}
\centering
\caption{ST prevalence across SAE architectures (Gemma-2-2B, L12).}
\label{tab:multiarch}
\vspace{-2mm}
\scriptsize
\begin{tabular}{@{}llrrr@{}}
\toprule
\textbf{SAE} & \textbf{Act.\ Fn} & \textbf{Width} & $n_\text{ST}$ & \textbf{\%ST} \\
\midrule
GemmaScope & JumpReLU & 16k & 158 & 0.96 \\
SAEBench & JumpReLU & 16k & 157 & 0.96 \\
SAEBench & TopK & 16k & 150 & 0.92 \\
SAEBench & BatchTopK & 16k & 156 & 0.95 \\
SAEBench & Mat.\ BTK & 16k & 162 & 0.99 \\
SAEBench & ReLU & 16k & 162 & 0.99 \\
SAEBench & Gated & 16k & 63 & 0.38 \\
SAEBench & P-Anneal & 16k & 68 & 0.42 \\
\midrule
GemmaScope & JumpReLU & 65k & 654 & 1.00 \\
Chanind & BTK$\to$JR & 32k & 274 & 0.84 \\
\bottomrule
\end{tabular}
\vspace{-4mm}
\end{wraptable}

To test whether activation function choice affects prevalence independently of training recipe, we compare eight SAE architectures trained on the same model (Gemma-2-2B) at Layer 12 using decoder-alignment detection (Table~\ref{tab:multiarch}).
Every 16k-width SAE yields similar prevalence (0.92--0.99\%)---only Gated (0.38\%) and P-Anneal (0.42\%) are lower.
The uniformity at mid-layers contrasts with Layer-0 differences---reinforcing that single-token features are primarily pre-compositional.

\subsection{Top-$k$ Sensitivity}
\label{app:k_sweep}

Detection uses $k = 20$ top activating tokens (Section~\ref{sec:methods}); this is the cache window exposed by the Neuronpedia public API for the original analysis pipeline.
To characterize sensitivity, we re-exported the top-10 cache and computed detection counts under smaller $k$ on the GPT2-Small cache, holding the gap and purity thresholds fixed at the canonical operating point: gap $\geq 0.3$, purity $\geq 0.6$, complete word.

\begin{table}[h]
\centering
\caption{Top-$k$ sensitivity on GPT2-Small (gap $\geq 0.3$, purity $\geq 0.6$, complete word).}
\label{tab:k_sweep}
\vspace{-2mm}
\small
\begin{tabular}{@{}ccl@{}}
\toprule
$k$ & $n_{\text{ST}}$ & \textbf{Trend} \\
\midrule
5 & 219 & re-exported cache \\
7 & 171 & $-$22\% from $k=5$ \\
10 & 164 & $-$4\% from $k=7$ \\
20 & 364 & original cache; main-text operating point \\
\bottomrule
\end{tabular}
\end{table}

Within the re-exported cache the count declines and then flattens (219 $\to$ 171 $\to$ 164, a 4\% step from $k = 7$ to $k = 10$): a feature with one strong primary activation and a long tail of weaker activations can satisfy the purity threshold at small $k$, but the additional cache positions surface enough secondary tokens to fail purity for some borderline features.
The $k = 20$ count comes from the original Neuronpedia export rather than this re-export, so it is not on a common scale with the sweep; what the sweep bounds is how many borderline features a given cache window admits, and the main-text results are computed at $k = 20$ throughout.

\subsection{Detection Threshold Ablation}

\begin{table}[h]
\centering
\caption{Extended ablation on detection thresholds showing precision-recall tradeoff.}
\label{tab:ablation}
\vspace{-2mm}
\small
\begin{tabular}{@{}ccc|cccc|cc@{}}
\toprule
\textbf{Gap} & \textbf{Purity} & \textbf{Word} & \textbf{N} & \textbf{Expl\%} & \textbf{Sim} & \textbf{ID} & \textbf{Prec} & \textbf{Rec} \\
\midrule
0.15 & 0.4 & No & 12,847 & 41\% & 1.8$\times$ & 8$\times$ & Low & High \\
0.20 & 0.5 & No & 7,171 & 52\% & 2.1$\times$ & 11$\times$ & Med & Med \\
0.25 & 0.55 & No & 5,412 & 58\% & 2.8$\times$ & 13$\times$ & Med & Med \\
0.30 & 0.6 & No & 4,227 & 61\% & 3.4$\times$ & 15$\times$ & Med & Low \\
0.30 & 0.6 & Yes & 364 & 64\% & 4.7$\times$ & 2$\times$ & High & Low \\
0.40 & 0.7 & Yes & 148 & 81\% & 5.2$\times$ & 18$\times$ & High & VLow \\
\bottomrule
\end{tabular}
\\[2pt]
\footnotesize{Sim = similarity ratio (ST/Poly); ID = intrinsic dimension ratio; Prec/Rec = qualitative precision-recall.}
\end{table}

\section{Additional Visualizations}
\label{app:figures}

\begin{figure*}[h]
    \centering
    \includegraphics[width=\textwidth]{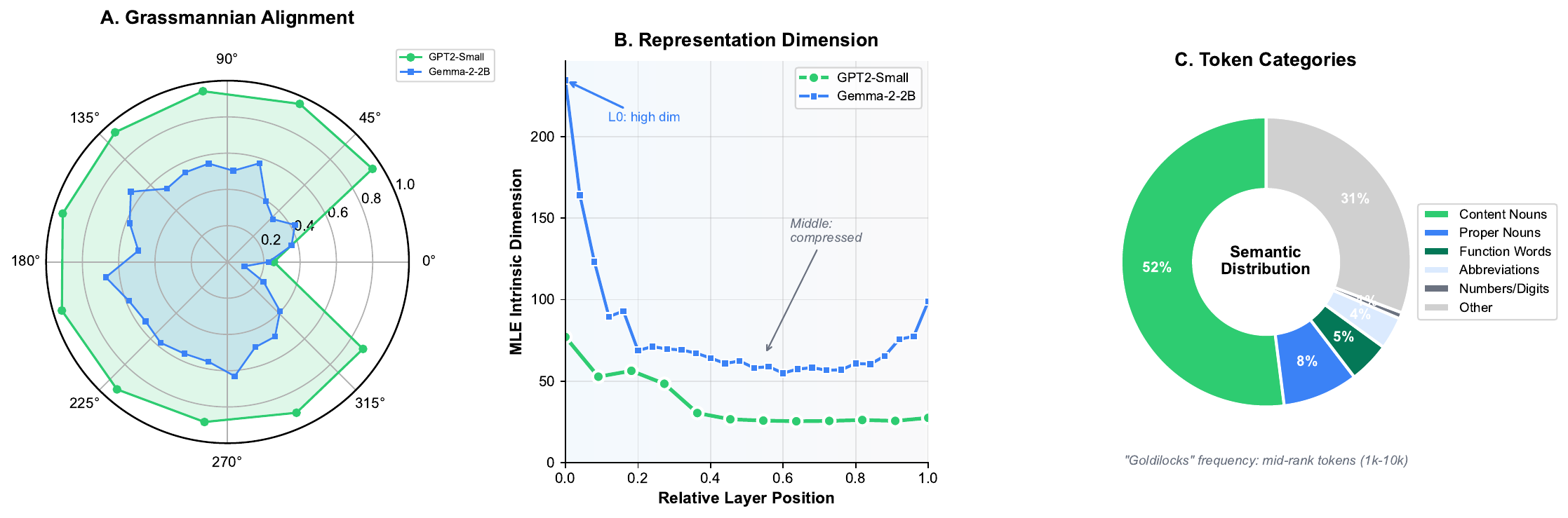}
    \caption{\textbf{Geometric analysis (mechanism overview).}
    (A) Grassmannian alignment for Gemma-2-2B: low L0$\to$L1 at 0.23, gradual increase through middle layers.
    (B) Intrinsic dimension: high at L0, compresses through middle layers, and rises again at final layers.
    (C) Semantic categories: content nouns at 52\% and proper nouns at 8.4\% dominate.}
    \label{fig:mechanism}
\end{figure*}

\subsection{Activation Shape Analysis}

\begin{figure}[ht]
    \centering
    \includegraphics[width=\textwidth]{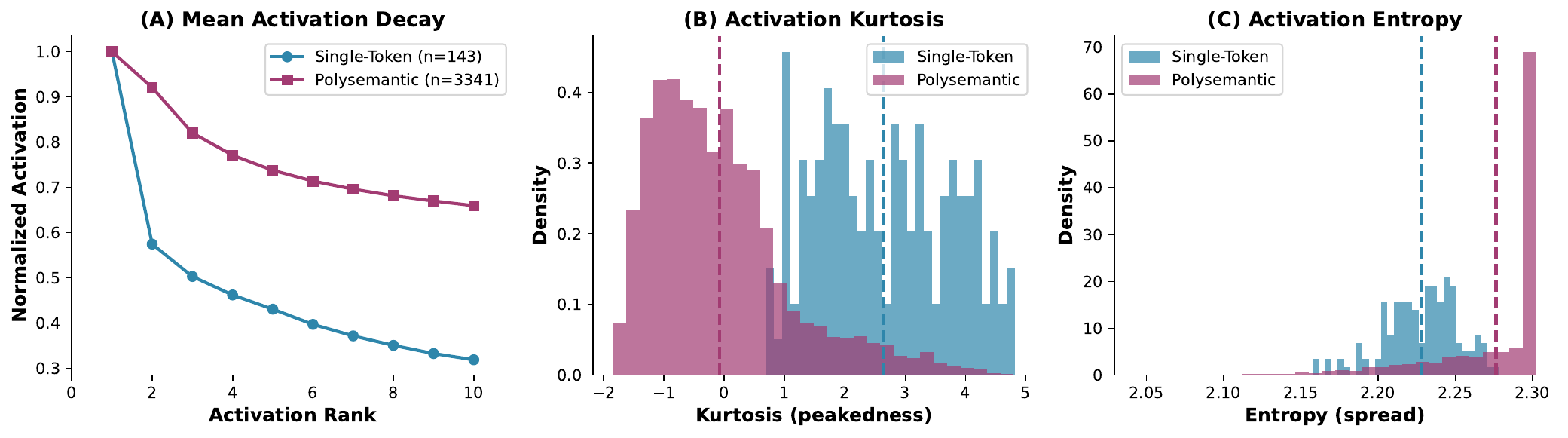}
    \caption{\textbf{Activation shape comparison.} (A) Mean activation decay across top-10 tokens: single-token features show sharp drop-off while polysemantic remain flat. (B) Kurtosis distribution: single-token features are sharply peaked (mean 2.65, $p < 10^{-154}$) vs polysemantic (mean $-$0.08). (C) Entropy distribution: single-token features show lower entropy (2.23 vs 2.28), confirming concentrated activations.}
    \label{fig:activation_shape}
\end{figure}

Single-token and polysemantic features differ most in their activation distributions across the vocabulary.
Figure~\ref{fig:activation_shape} shows three complementary measures of activation shape.
Panel (A) shows how normalized activation values decay across the top-10 activating tokens: single-token features exhibit a sharp drop-off (from 1.0 to $\sim$0.3 by rank 10), while polysemantic features maintain relatively flat activation profiles (from 1.0 to $\sim$0.65): single-token features respond strongly to one token and weakly to others; polysemantic features respond moderately to many tokens.

Panel (B) shows the distribution of kurtosis (peakedness) across feature types.
Single-token features have mean kurtosis 2.65, indicating sharply peaked distributions with heavy tails---while polysemantic features have mean kurtosis $-$0.08, indicating flat, uniform-like distributions.
This difference is highly significant ($p < 10^{-154}$) and provides a statistical signature independent of our gap ratio metric.
Panel (C) shows entropy distributions: single-token features have lower entropy (mean 2.23) than polysemantic (mean 2.28), confirming their more concentrated activation patterns.
Together, these metrics validate that our detection method captures features with genuinely distinct activation behavior.

\subsection{Token Frequency Distribution}

\begin{figure}[ht]
    \centering
    \includegraphics[width=0.75\textwidth]{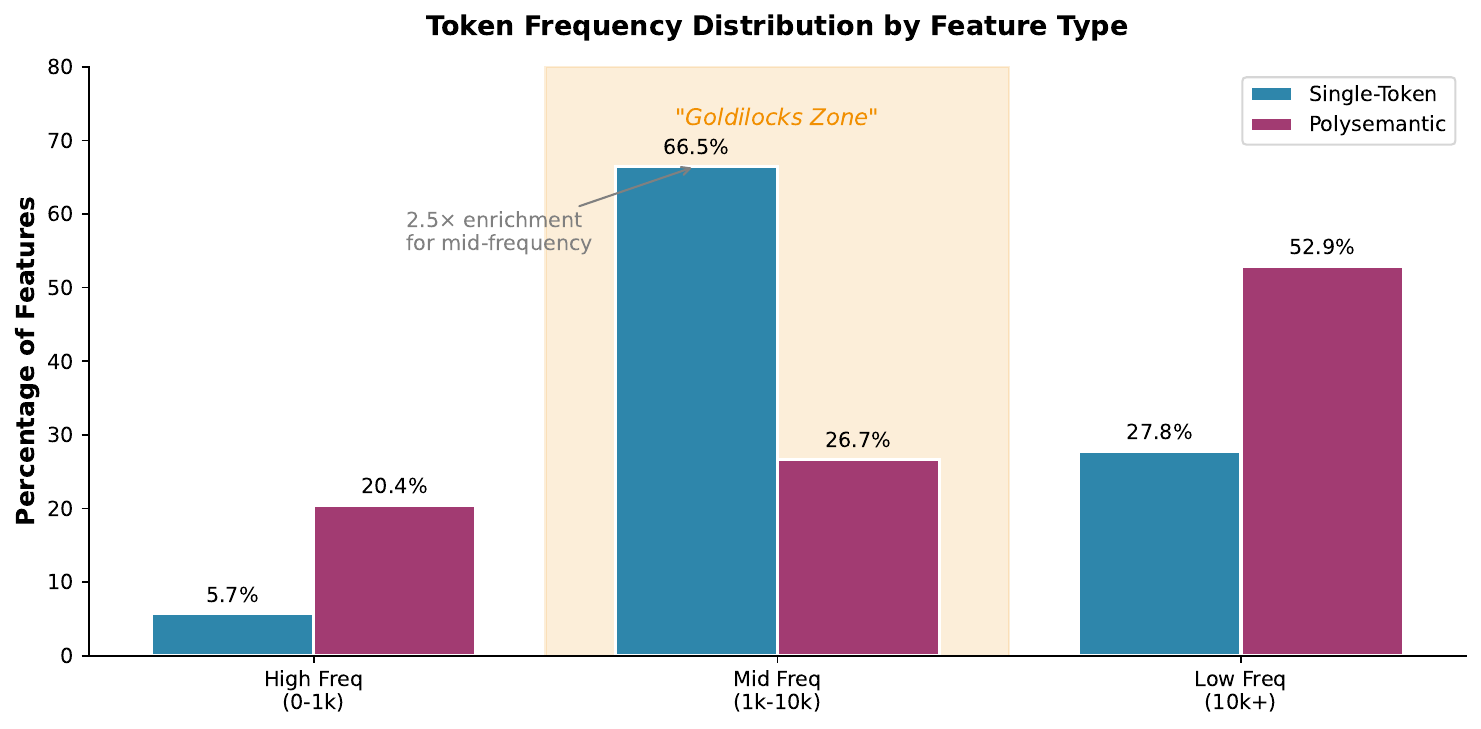}
    \caption{\textbf{Token frequency distribution by feature type.}
    Single-token features show 2.5$\times$ enrichment for mid-frequency tokens (rank 1k--10k), avoiding both high-frequency function words that require compositional encoding and rare tokens that must share capacity.
    This ``Goldilocks zone'' pattern suggests single-token features encode tokens that benefit from dedicated, context-independent representations.}
    \label{fig:goldilocks}
\end{figure}

Why do single-token features encode some tokens but not others?
Figure~\ref{fig:goldilocks} shows the encoding pattern.
Single-token features preferentially encode mid-frequency tokens (the ``Goldilocks zone''): in GPT2-Small Layer 0, 66.5\% of single-token features correspond to tokens ranked 1k--10k in frequency, compared to 26.7\% for polysemantic features, a 2.5$\times$ enrichment.
Conversely, high-frequency function words (rank $<$1k) comprise only 5.7\% of single-token features versus 20.4\% of polysemantic, and low-frequency tokens (rank $>$10k) are also underrepresented (27.8\% vs 52.9\%).

Mid-frequency tokens are common enough to warrant dedicated neural representations, unlike rare tokens that must share capacity, yet not so frequent that the model can rely on compositional or contextual features as it does for function words.
Content nouns, proper nouns, and function words dominate the single-token population (Table~\ref{tab:semantic}), categories that benefit from stable, context-independent representations.

\subsection{Cross-Layer Persistence by Feature Type}

\begin{figure}[ht]
    \centering
    \includegraphics[width=0.75\textwidth]{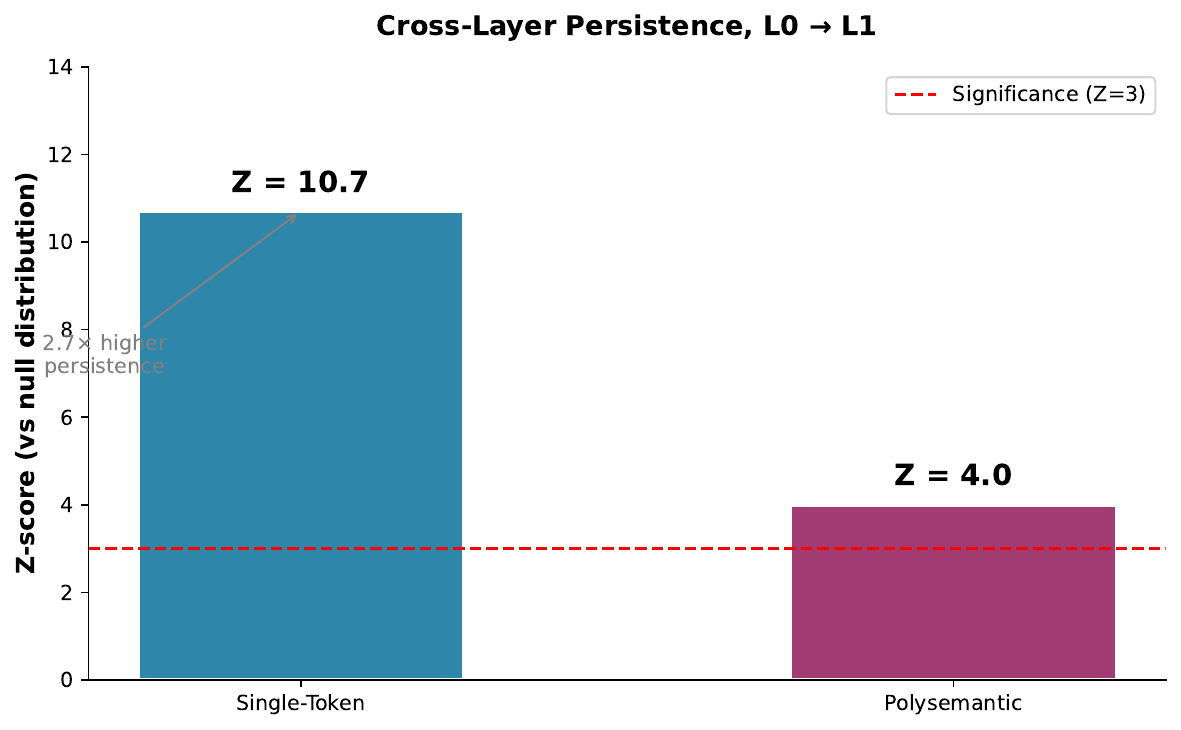}
    \caption{\textbf{Cross-layer persistence by feature type.} Single-token features show Z-score 10.7 (vs null distribution of random pairs), 2.7$\times$ higher than polysemantic features (Z=4.0, $p < 10^{-73}$). The dashed line indicates the significance threshold (Z=3). This confirms single-token features serve as stable reference points that persist through the L0$\rightarrow$L1 representational shift.}
    \label{fig:persistence}
\end{figure}

Do single-token features persist more across layers than polysemantic features?
Figure~\ref{fig:persistence} quantifies this by computing the Z-score of each feature's cross-layer similarity against a null distribution of random vector pairs.
Single-token features show Z-score 10.7 for L0$\to$L1 persistence, meaning their cross-layer similarity is 10.7 standard deviations above what would be expected by chance.
Polysemantic features show Z-score 4.0, still significant, but 2.7$\times$ lower than single-token features ($p < 10^{-73}$).

This difference is consistent with single-token features maintaining stable directions across layers because they encode fixed token identities, while polysemantic features encode context-dependent combinations and must transform more as contextual information accumulates.
Both feature types undergo the same L0$\to$L1 shift, but single-token features are more likely to survive it with their direction intact.

\subsection{Semantic Preservation in Persistent Features}

\begin{figure}[ht]
    \centering
    \includegraphics[width=0.65\textwidth]{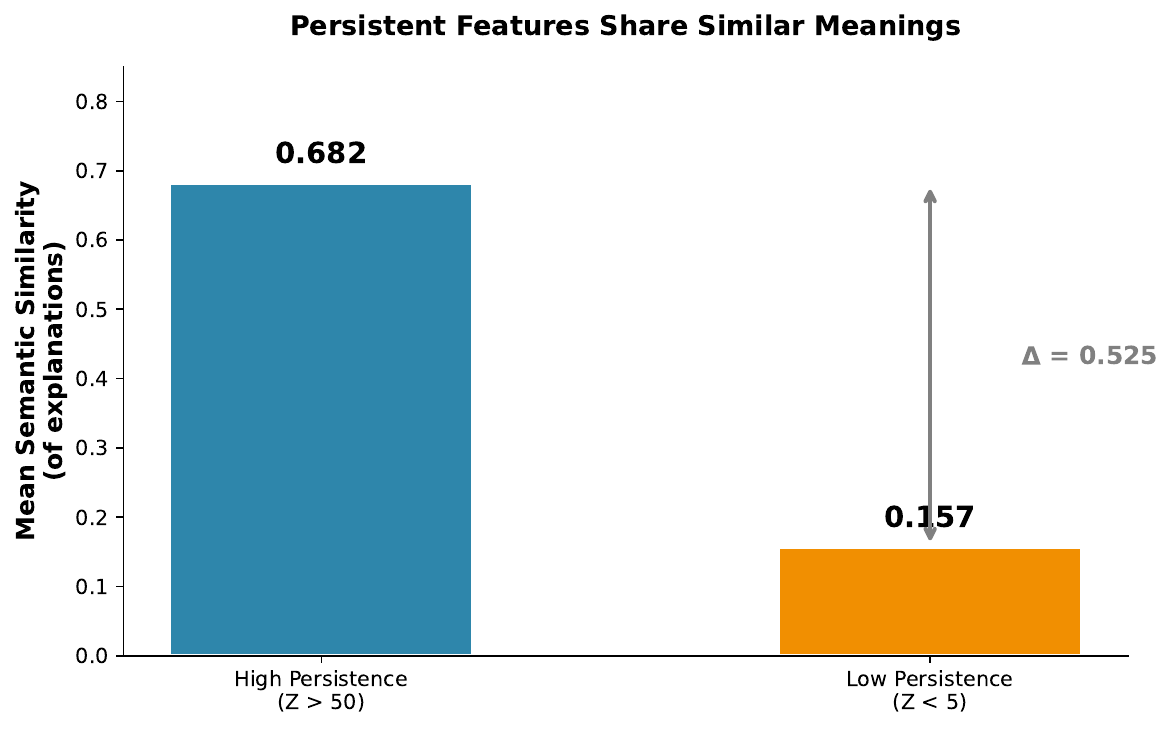}
    \caption{\textbf{Semantic preservation in persistent features.} High-persistence feature pairs (Z $>$ 50) show explanation similarity 0.682, while low-persistence pairs (Z $<$ 5) show only 0.157, a difference of 0.525. This confirms that geometric persistence corresponds to genuine semantic continuity: features that maintain similar directions across layers also maintain similar meanings.}
    \label{fig:semantic}
\end{figure}

Persistence alone does not guarantee semantic preservation---a feature could maintain a similar direction while encoding entirely different concepts at each layer.
Figure~\ref{fig:semantic} tests whether persistent features really share semantic meaning: it compares the auto-generated explanations of matched feature pairs across layers.
We measure semantic similarity with sentence embeddings of the Neuronpedia explanations.

High-persistence feature pairs at $Z > 50$, indicating very strong cross-layer similarity, show mean explanation similarity of 0.682, while low-persistence pairs at $Z < 5$ show only 0.157, a difference of 0.525.
Manual inspection confirms it---high-persistence pairs are overwhelmingly single-token features encoding proper names (``Cruz,'' ``Matt,'' ``Scott''), where the L5 and L6 explanations both reference the same token.
Our geometric persistence metric therefore captures genuine semantic continuity, not mere directional accident.
The pattern replicates in Gemma-2-2B at 0.61 vs 0.32, difference 0.29: a cross-model phenomenon.

Despite different tokenizers, GPT2 and Gemma single-token features show semantic correspondence.
Table~\ref{tab:correspondence} breaks down match rates by semantic category: proper nouns show highest cross-model correspondence (33\%) and function words lowest (14\%).
Partial semantic convergence despite zero surface-form overlap.

\begin{table}[h]
\centering
\caption{Cross-model feature correspondence by semantic category.}
\label{tab:correspondence}
\small
\begin{tabular}{@{}l|ccc|cc@{}}
\toprule
\textbf{Category} & \textbf{GPT2} & \textbf{Matched} & \textbf{Rate} & \textbf{Avg Sim} & \textbf{Max Sim} \\
\midrule
Proper Nouns & 138 & 45 & 33\% & 0.78 & 0.94 \\
Common Verbs & 87 & 18 & 21\% & 0.74 & 0.89 \\
Concrete Nouns & 76 & 15 & 20\% & 0.72 & 0.87 \\
Adjectives & 41 & 6 & 15\% & 0.71 & 0.83 \\
Function Words & 22 & 3 & 14\% & 0.68 & 0.76 \\
\midrule
\textbf{Total} & 364 & 87 & 24\% & 0.75 & 0.94 \\
\bottomrule
\end{tabular}
\end{table}

\subsection{Decoder-Embedding Alignment}

\begin{figure}[ht]
    \centering
    \includegraphics[width=0.65\textwidth]{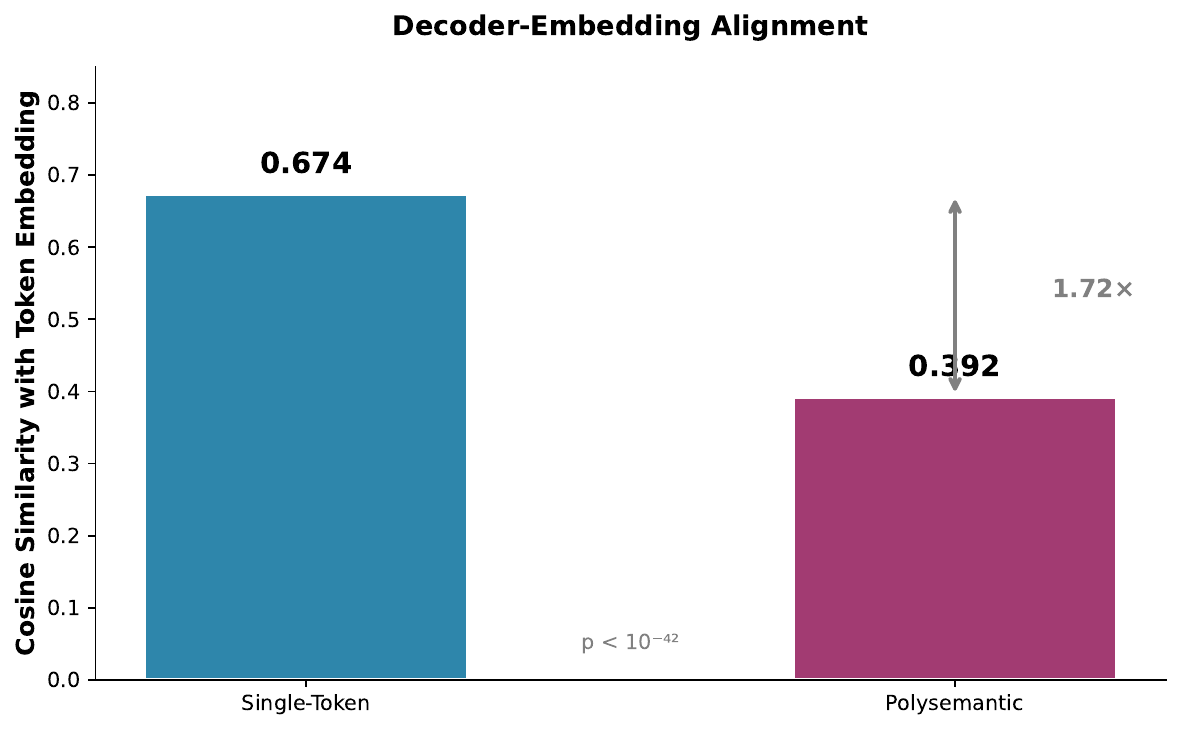}
    \caption{\textbf{Decoder-embedding alignment.} Single-token decoder vectors show 1.72$\times$ higher cosine similarity with corresponding token embeddings (0.674 vs 0.392, $p < 10^{-42}$). This mechanistic validation confirms single-token features recover directions close to the original embedding space, functioning as stable reference points for token identity.}
    \label{fig:alignment}
\end{figure}

If single-token features truly encode token identity, their decoder vectors should align with the corresponding token embeddings.
Figure~\ref{fig:alignment} tests this prediction by computing cosine similarity between each feature's decoder vector $\mathbf{w}_\text{dec}$ and the embedding of its top activating token $\mathbf{e}_\text{tok}$.
Single-token features show mean alignment 0.674 versus 0.392 for polysemantic, a 1.72$\times$ difference; $t = 14.4$, $p < 10^{-42}$.

This alignment is independent of activation patterns: single-token decoder vectors point toward the token embedding subspace, close to the original input representation.
The alignment decreases with layer depth (not shown), consistent with representations becoming more abstract in later layers.
Combined with the logit attribution analysis, where 89\% of single-token features have the top-token match, this confirms single-token features function as bidirectional bridges between the embedding and unembedding spaces.

\subsection{Full-Layer Causal Ablation Results}
\label{app:full_layer}

Tables~\ref{tab:full_gemma2}--\ref{tab:full_gemma3} present layer-wise necessity $p$-values for the four full-layer experiments summarized in Table~\ref{tab:causal}.
All $p$-values are from Mann-Whitney $U$ tests of ST vs size-matched random controls.
$\ast$: BH-corrected $p < 0.05$.

\begin{table}[h]
\centering
\begin{minipage}[t]{0.48\textwidth}
\centering
\caption{Gemma-2-2B: GemmaScope (26/26 sig) vs BatchTopK (24/25 sig).}
\label{tab:full_gemma2}
\vspace{-2mm}
\tiny
\begin{tabular}{@{}r|rc|rc@{}}
\toprule
& \multicolumn{2}{c|}{\textbf{GemmaScope (JR)}} & \multicolumn{2}{c}{\textbf{BatchTopK}} \\
\textbf{L} & $n$ & \textbf{$p$} & $n$ & \textbf{$p$} \\
\midrule
0 & 82 & $2.5\mathrm{e}{-4}$$\ast$ & 260 & $0.028$$\ast$ \\
1 & 156 & $0.021$$\ast$ & 64 & $0.047$ \\
2 & 157 & $2.9\mathrm{e}{-4}$$\ast$ & 164 & $0.036$$\ast$ \\
3 & 149 & $4.4\mathrm{e}{-5}$$\ast$ & 188 & $0.040$$\ast$ \\
4 & 141 & $0.004$$\ast$ & 211 & $0.009$$\ast$ \\
5 & 134 & $2.1\mathrm{e}{-4}$$\ast$ & 298 & $9.4\mathrm{e}{-11}$$\ast$ \\
6 & 131 & $0.002$$\ast$ & 322 & $5.4\mathrm{e}{-33}$$\ast$ \\
7 & 121 & $4.3\mathrm{e}{-4}$$\ast$ & 316 & $1.1\mathrm{e}{-48}$$\ast$ \\
8 & 118 & $0.001$$\ast$ & 317 & $1.6\mathrm{e}{-62}$$\ast$ \\
9 & 99 & $0.004$$\ast$ & 318 & $\mathbf{2.9\mathrm{e}{-67}}$$\ast$ \\
10 & 103 & $1.0\mathrm{e}{-4}$$\ast$ & 242 & $1.2\mathrm{e}{-25}$$\ast$ \\
11 & 110 & $3.2\mathrm{e}{-5}$$\ast$ & 225 & $2.0\mathrm{e}{-36}$$\ast$ \\
12 & 129 & $0.008$$\ast$ & 227 & $1.0\mathrm{e}{-43}$$\ast$ \\
13 & 97 & $1.8\mathrm{e}{-4}$$\ast$ & 216 & $2.1\mathrm{e}{-33}$$\ast$ \\
14 & 97 & $0.001$$\ast$ & 73 & $5.1\mathrm{e}{-9}$$\ast$ \\
15 & 93 & $0.002$$\ast$ & 151 & $3.8\mathrm{e}{-20}$$\ast$ \\
16 & 68 & $0.021$$\ast$ & 122 & $5.4\mathrm{e}{-21}$$\ast$ \\
17 & 63 & $7.6\mathrm{e}{-5}$$\ast$ & 71 & $4.3\mathrm{e}{-11}$$\ast$ \\
18 & 60 & $1.3\mathrm{e}{-6}$$\ast$ & 82 & $9.2\mathrm{e}{-18}$$\ast$ \\
19 & 51 & $3.3\mathrm{e}{-6}$$\ast$ & 99 & $4.1\mathrm{e}{-21}$$\ast$ \\
20 & 55 & $1.1\mathrm{e}{-9}$$\ast$ & 83 & $5.1\mathrm{e}{-16}$$\ast$ \\
21 & 65 & $6.9\mathrm{e}{-13}$$\ast$ & 97 & $2.9\mathrm{e}{-12}$$\ast$ \\
22 & 89 & $9.4\mathrm{e}{-15}$$\ast$ & 124 & $1.5\mathrm{e}{-21}$$\ast$ \\
23 & 80 & $3.8\mathrm{e}{-15}$$\ast$ & 138 & $1.2\mathrm{e}{-30}$$\ast$ \\
24 & 77 & $7.0\mathrm{e}{-16}$$\ast$ & 166 & $1.3\mathrm{e}{-23}$$\ast$ \\
25 & 101 & $1.4\mathrm{e}{-34}$$\ast$ & \multicolumn{2}{c}{} \\
\bottomrule
\end{tabular}
\\[2pt]
{\scriptsize GemmaScope peaks at late layers (L20+); BatchTopK peaks at mid-layers (L6--L9).}
\end{minipage}
\hfill
\begin{minipage}[t]{0.48\textwidth}
\centering
\caption{Gemma-3-1B: GemmaScope (9/26 sig) vs BatchTopK (18/25 sig).}
\label{tab:full_gemma3}
\vspace{-2mm}
\tiny
\begin{tabular}{@{}r|rc|rc@{}}
\toprule
& \multicolumn{2}{c|}{\textbf{GemmaScope (JR)}} & \multicolumn{2}{c}{\textbf{BatchTopK}} \\
\textbf{L} & $n$ & \textbf{$p$} & $n$ & \textbf{$p$} \\
\midrule
0 & 89 & $0.100$ & 236 & $0.062$ \\
1 & 59 & $0.568$ & 88 & $0.585$ \\
2 & 56 & $0.103$ & 83 & $0.335$ \\
3 & 64 & $0.095$ & 50 & $0.096$ \\
4 & 109 & $0.149$ & 33 & $0.029$$\ast$ \\
5 & 142 & $0.006$$\ast$ & 25 & $0.014$$\ast$ \\
6 & 141 & $0.001$$\ast$ & 29 & $5.6\mathrm{e}{-4}$$\ast$ \\
7 & 138 & $0.505$ & 27 & $0.590$ \\
8 & 150 & $0.233$ & 32 & $0.026$$\ast$ \\
9 & 141 & $0.140$ & 37 & $0.085$ \\
10 & 148 & $0.708$ & 30 & $0.167$ \\
11 & 144 & $0.485$ & 50 & $9.5\mathrm{e}{-12}$$\ast$ \\
12 & 144 & $0.164$ & 60 & $3.4\mathrm{e}{-8}$$\ast$ \\
13 & 142 & $0.266$ & 50 & $1.4\mathrm{e}{-12}$$\ast$ \\
14 & 136 & $0.037$$\ast$ & 61 & $4.7\mathrm{e}{-14}$$\ast$ \\
15 & 139 & $0.117$ & 61 & $1.7\mathrm{e}{-17}$$\ast$ \\
16 & 132 & $0.001$$\ast$ & 96 & $3.8\mathrm{e}{-24}$$\ast$ \\
17 & 128 & $0.005$$\ast$ & 141 & $8.3\mathrm{e}{-25}$$\ast$ \\
18 & 128 & $0.022$$\ast$ & 149 & $4.5\mathrm{e}{-27}$$\ast$ \\
19 & 128 & $0.157$ & 148 & $7.2\mathrm{e}{-19}$$\ast$ \\
20 & 114 & $0.319$ & 153 & $2.7\mathrm{e}{-29}$$\ast$ \\
21 & 127 & $0.003$$\ast$ & 151 & $9.2\mathrm{e}{-26}$$\ast$ \\
22 & 116 & $0.778$ & 148 & $1.5\mathrm{e}{-25}$$\ast$ \\
23 & 137 & $7.8\mathrm{e}{-8}$$\ast$ & 166 & $1.1\mathrm{e}{-29}$$\ast$ \\
24 & 135 & $0.001$$\ast$ & 161 & $7.9\mathrm{e}{-29}$$\ast$ \\
25 & 119 & $0.610$ & \multicolumn{2}{c}{} \\
\bottomrule
\end{tabular}
\\[2pt]
{\scriptsize BatchTopK ST count inverts: peaks at late layers (L17--L24) vs GemmaScope early-mid.}
\end{minipage}
\end{table}

\subsection{Full-Layer Extension: Llama-3.1-8B (G1) and Gemma-2-9B (G2)}
\label{app:full_layer_extension}

Tables~\ref{tab:full_llama8b} and~\ref{tab:full_gemma9b} report layer-wise necessity ($p_{\text{BH}}$ and mean $\Delta$logit) for two further full-depth experiments: Llama-3.1-8B $\times$ LlamaScope (32 layers) and Gemma-2-9B $\times$ GemmaScope (42 layers).
All $p$-values are BH-corrected globally across all layers in each configuration; $\ast$: $p_{\text{BH}} < 0.05$.

\begin{table}[h]
\centering
\begin{minipage}[t]{0.48\textwidth}
\centering
\caption{Llama-3.1-8B $\times$ LlamaScope, full 32 layers (G1). 31/32 layers BH-significant for necessity; peak $p_{\text{BH}} = 2.3 \times 10^{-141}$ at L1. Anchor: 10/32 source layers show $\geq 1$ significantly disrupted downstream layer.}
\label{tab:full_llama8b}
\vspace{-2mm}
\tiny
\begin{tabular}{@{}r|rcr@{}}
\toprule
\textbf{L} & $n_{\text{ST}}$ & \textbf{$p_{\text{BH}}$} & $\Delta$logit \\
\midrule
0 & 280 & $8.1\mathrm{e}{-6}$$\ast$ & $-0.006$ \\
1 & 328 & $\mathbf{2.3\mathrm{e}{-141}}$$\ast$ & $-1.828$ \\
2 & 295 & $4.0\mathrm{e}{-98}$$\ast$ & $-0.701$ \\
3 & 292 & $1.4\mathrm{e}{-98}$$\ast$ & $-0.847$ \\
4 & 290 & $3.0\mathrm{e}{-89}$$\ast$ & $-0.741$ \\
5 & 248 & $4.9\mathrm{e}{-65}$$\ast$ & $-0.585$ \\
6 & 141 & $4.4\mathrm{e}{-34}$$\ast$ & $-0.565$ \\
7 & 81 & $1.5\mathrm{e}{-24}$$\ast$ & $-0.795$ \\
8 & 53 & $5.6\mathrm{e}{-11}$$\ast$ & $-0.510$ \\
9 & 42 & $2.5\mathrm{e}{-8}$$\ast$ & $-0.671$ \\
10 & 28 & $1.1\mathrm{e}{-10}$$\ast$ & $-0.454$ \\
11 & 36 & $2.5\mathrm{e}{-7}$$\ast$ & $-0.615$ \\
12 & 37 & $1.4\mathrm{e}{-8}$$\ast$ & $-0.647$ \\
13 & 48 & $4.1\mathrm{e}{-14}$$\ast$ & $-0.925$ \\
14 & 55 & $3.8\mathrm{e}{-10}$$\ast$ & $-0.847$ \\
15 & 64 & $3.7\mathrm{e}{-17}$$\ast$ & $-0.807$ \\
16 & 54 & $3.7\mathrm{e}{-17}$$\ast$ & $-0.988$ \\
17 & 55 & $8.1\mathrm{e}{-14}$$\ast$ & $-0.936$ \\
18 & 46 & $1.8\mathrm{e}{-12}$$\ast$ & $-0.854$ \\
19 & 37 & $2.0\mathrm{e}{-9}$$\ast$ & $-0.642$ \\
20 & 39 & $7.2\mathrm{e}{-13}$$\ast$ & $-0.710$ \\
21 & 38 & $1.5\mathrm{e}{-11}$$\ast$ & $-0.560$ \\
22 & 35 & $6.9\mathrm{e}{-8}$$\ast$ & $-0.443$ \\
23 & 39 & $6.9\mathrm{e}{-8}$$\ast$ & $-0.360$ \\
24 & 36 & $5.8\mathrm{e}{-9}$$\ast$ & $-0.414$ \\
25 & 34 & $2.8\mathrm{e}{-12}$$\ast$ & $-0.559$ \\
26 & 32 & $7.0\mathrm{e}{-10}$$\ast$ & $-0.284$ \\
27 & 36 & $5.6\mathrm{e}{-16}$$\ast$ & $-0.546$ \\
28 & 33 & $9.6\mathrm{e}{-12}$$\ast$ & $-0.354$ \\
29 & 26 & $6.9\mathrm{e}{-8}$$\ast$ & $-0.276$ \\
30 & 13 & $2.2\mathrm{e}{-4}$$\ast$ & $-0.108$ \\
31 & 1 & $5.0\mathrm{e}{-1}$ & $0.000$ \\
\bottomrule
\end{tabular}
\\[2pt]
{\scriptsize Last layer (L31) has $n_{\text{ST}} = 1$ feature, insufficient for significance.}
\end{minipage}
\hfill
\begin{minipage}[t]{0.48\textwidth}
\centering
\caption{Gemma-2-9B $\times$ GemmaScope, full 42 layers (G2). 42/42 BH-significant; peak $p_{\text{BH}} = 6.1\mathrm{e}{-26}$ at L35; anchor 41/42; Spearman $\rho(\text{depth}, |\Delta\text{logit}|) = 0.81$ ($p<10^{-9}$).}
\label{tab:full_gemma9b}
\vspace{-2mm}
\tiny
\begin{tabular}{@{}r|rcr@{}}
\toprule
\textbf{L} & $n_{\text{ST}}$ & \textbf{$p_{\text{BH}}$} & $\Delta$logit \\
\midrule
0 & 76 & $1.6\mathrm{e}{-6}$$\ast$ & $-0.182$ \\
1 & 105 & $4.2\mathrm{e}{-8}$$\ast$ & $-0.230$ \\
2 & 134 & $4.8\mathrm{e}{-9}$$\ast$ & $-0.037$ \\
3 & 133 & $1.3\mathrm{e}{-6}$$\ast$ & $-0.042$ \\
4 & 157 & $2.9\mathrm{e}{-5}$$\ast$ & $-0.010$ \\
5 & 150 & $2.1\mathrm{e}{-7}$$\ast$ & $-0.012$ \\
6 & 151 & $9.1\mathrm{e}{-11}$$\ast$ & $-0.015$ \\
7 & 145 & $3.5\mathrm{e}{-3}$$\ast$ & $-0.012$ \\
8 & 141 & $3.1\mathrm{e}{-4}$$\ast$ & $-0.006$ \\
9 & 123 & $1.1\mathrm{e}{-5}$$\ast$ & $-0.033$ \\
10 & 129 & $5.7\mathrm{e}{-6}$$\ast$ & $-0.009$ \\
11 & 112 & $4.9\mathrm{e}{-5}$$\ast$ & $-0.018$ \\
12 & 103 & $6.3\mathrm{e}{-3}$$\ast$ & $-0.010$ \\
13 & 101 & $1.8\mathrm{e}{-3}$$\ast$ & $-0.017$ \\
14 & 76 & $7.3\mathrm{e}{-5}$$\ast$ & $-0.027$ \\
15 & 85 & $2.6\mathrm{e}{-4}$$\ast$ & $-0.036$ \\
16 & 66 & $1.2\mathrm{e}{-2}$$\ast$ & $-0.043$ \\
17 & 62 & $2.4\mathrm{e}{-3}$$\ast$ & $-0.094$ \\
18 & 58 & $2.2\mathrm{e}{-5}$$\ast$ & $-0.124$ \\
19 & 61 & $5.6\mathrm{e}{-8}$$\ast$ & $-0.103$ \\
20 & 53 & $1.0\mathrm{e}{-6}$$\ast$ & $-0.209$ \\
21 & 53 & $1.6\mathrm{e}{-7}$$\ast$ & $-0.193$ \\
22 & 65 & $1.9\mathrm{e}{-8}$$\ast$ & $-0.199$ \\
23 & 57 & $4.3\mathrm{e}{-9}$$\ast$ & $-0.242$ \\
24 & 47 & $2.7\mathrm{e}{-8}$$\ast$ & $-0.294$ \\
25 & 42 & $2.3\mathrm{e}{-11}$$\ast$ & $-0.416$ \\
26 & 42 & $1.1\mathrm{e}{-13}$$\ast$ & $-0.249$ \\
27 & 49 & $7.6\mathrm{e}{-14}$$\ast$ & $-0.221$ \\
28 & 56 & $1.8\mathrm{e}{-7}$$\ast$ & $-0.175$ \\
29 & 54 & $1.1\mathrm{e}{-13}$$\ast$ & $-0.279$ \\
30 & 69 & $2.0\mathrm{e}{-19}$$\ast$ & $-0.291$ \\
31 & 76 & $1.3\mathrm{e}{-22}$$\ast$ & $-0.313$ \\
32 & 77 & $1.2\mathrm{e}{-17}$$\ast$ & $-0.273$ \\
33 & 75 & $5.8\mathrm{e}{-20}$$\ast$ & $-0.240$ \\
34 & 77 & $1.3\mathrm{e}{-22}$$\ast$ & $-0.268$ \\
35 & 82 & $\mathbf{6.1\mathrm{e}{-26}}$$\ast$ & $-0.333$ \\
36 & 83 & $3.8\mathrm{e}{-22}$$\ast$ & $-0.343$ \\
37 & 92 & $5.2\mathrm{e}{-24}$$\ast$ & $-0.340$ \\
38 & 84 & $3.7\mathrm{e}{-21}$$\ast$ & $-0.441$ \\
39 & 98 & $6.1\mathrm{e}{-26}$$\ast$ & $-0.616$ \\
40 & 119 & $2.8\mathrm{e}{-21}$$\ast$ & $-0.403$ \\
41 & 141 & $2.4\mathrm{e}{-23}$$\ast$ & $-0.219$ \\
\bottomrule
\end{tabular}
\end{minipage}
\end{table}

\subsection{Paired Comparison: BatchTopK vs GemmaScope}
\label{app:paired}

To test whether the BatchTopK advantage reflects a genuine budget pressure effect, we perform a token-ID matched paired comparison between GemmaScope and BatchTopK on two models (Gemma-2-2B and Gemma-3-1B).
For each layer, we identify tokens detected as single-token features by \textit{both} SAE types, yielding $N\!=\!627$ matched pairs---473 from Gemma-2-2B L0--L24 and 154 from Gemma-3-1B.

\begin{table}[h]
\centering
\caption{Paired comparison: BatchTopK vs GemmaScope (token-ID matched, $N\!=\!627$).}
\label{tab:paired}
\vspace{-2mm}
\small
\begin{tabular}{@{}lccccc@{}}
\toprule
\textbf{Metric} & \textbf{BatchTopK} & \textbf{GemmaScope} & \textbf{$p$} & \textbf{$r$} & \textbf{Direction} \\
\midrule
Anchor ($\Delta$logit-lens) & 15{,}537 & 8{,}554 & $1.2\!\times\!10^{-18}$ & 0.36 & BTK $>$ GS \\
Necessity ($\Delta$rank) & 277.7 & 195.7 & 0.004 & 0.12 & BTK $>$ GS \\
\quad L0--L18 only & 323.8 & 111.9 & $2.0\!\times\!10^{-6}$ & 0.28 & BTK $>$ GS \\
Recovery (rank within 2$\times$) & 67.5\% & 65.6\% & 0.41 & 0.03 & ns \\
\bottomrule
\end{tabular}
\end{table}

BatchTopK features show stronger downstream anchoring ($p = 1.2 \times 10^{-18}$, $r\!=\!0.36$) and stronger necessity ($p\!=\!0.004$).
The effect is layer-dependent: early-mid layers (L0--L18) show stronger necessity ($p = 2.0 \times 10^{-6}$), while late layers converge---consistent with output-proximal layers enforcing token identity regardless of activation function.

\subsection{Controlled Comparison: TopK vs JumpReLU}
\label{app:controlled}

\begin{table}[h]
\centering
\caption{Controlled comparison: TopK vs JumpReLU ($N\!=\!142$ token-matched).}
\label{tab:controlled}
\vspace{-2mm}
\small
\begin{tabular}{@{}lccccc@{}}
\toprule
\textbf{Metric} & \textbf{TopK} & \textbf{JumpReLU} & \textbf{$p$} & \textbf{$r$} & \textbf{Direction} \\
\midrule
Necessity (mean $\Delta$rank) & 199.5 & 311.0 & 0.036 & 0.20 & JR $>$ TopK \\
Necessity (max $\Delta$rank) & 325.5 & 513.4 & 0.003 & 0.27 & JR $>$ TopK \\
\bottomrule
\end{tabular}
\end{table}

To disentangle activation function from training recipe, we compare community SAEs on Gemma-2-2B L1 with the same dictionary width (18k): TopK ($k \in \{25, 50, 75, 100\}$) and JumpReLU ($\ell_0 \in \{14, 26, 95\}$), all using tied decoders without norm constraint.
The controlled comparison (Table~\ref{tab:controlled}) shows the \textit{opposite} direction from the cross-model result (Table~\ref{tab:paired}): JumpReLU features cause more damage than TopK ($p\!=\!0.036$, matched-pair rank-biserial $r\!=\!0.20$).
This dissociation indicates the activation function alone does not explain the cross-SAE-family differences.
Training recipe factors such as decoder norms, training scale, and post-hoc conversion remain the residual candidates at the recipe level.
Comparing the community JumpReLU SAE with GemmaScope L1, both JumpReLU but differing in recipe, shows lower degradation for GemmaScope at median $0.51$ vs $2.06$ $\log_2$ with $p = 3.1 \times 10^{-15}$, consistent with unit-norm decoders reducing per-feature perturbation.

\subsection{Sparsity Dose-Response Analysis}
\label{app:dose_response}

\begin{table}[h]
\centering
\caption{Dose-response: mean $\Delta$logit by sparsity level.
Tighter TopK budget monotonically increases necessity; JumpReLU shows no trend.}
\label{tab:dose_response}
\vspace{-2mm}
\small
\begin{tabular}{@{}llcc@{}}
\toprule
\textbf{SAE Type} & \textbf{$\ell_0$} & \textbf{Mean $\Delta$logit} & \textbf{$n_{\text{ST}}$} \\
\midrule
TopK & 25 & \hlbest{$-0.152$} & 169 \\
TopK & 50 & \hlbest{$-0.094$} & 176 \\
TopK & 75 & \hlbest{$-0.060$} & 180 \\
TopK & 100 & \hlbest{$-0.036$} & 179 \\
\midrule
JumpReLU & 14 & $-0.041$ & 175 \\
JumpReLU & 26 & $-0.039$ & 172 \\
JumpReLU & 95 & $-0.043$ & 170 \\
\bottomrule
\end{tabular}
\end{table}

To test whether competitive budget size monotonically predicts necessity, we test TopK SAEs at four sparsity levels ($k \in \{25, 50, 75, 100\}$) and JumpReLU SAEs at three levels ($\ell_0 \in \{14, 26, 95\}$), all on Gemma-2-2B L1---same dictionary width (18k) throughout: TopK shows a strong monotonic trend (Table~\ref{tab:dose_response}): Pearson $r\!=\!0.98$ ($p\!=\!0.020$, group-level) and Spearman $\rho\!=\!0.077$ ($p\!=\!0.04$, per-feature). $\Delta$rank shows the same monotonic pattern ($r\!=\!-0.928$, $p\!=\!0.072$).
JumpReLU shows no dose-response ($r\!=\!-0.015$, ns), consistent with its variable budget---changing the threshold does not create competitive allocation pressure.

\subsection{Cross-SAE Token-Matched Analysis}
\label{app:cross_sae}

To understand which tokens are robust versus sensitive to SAE methodology, we categorize the $N\!=\!474$ token-matched pairs (Gemma-2-2B, GemmaScope vs BatchTopK) into semantic categories and compare convergence and anchoring dominance (Table~\ref{tab:per_token_category}).

\begin{table}[h]
\centering
\caption{SAE methodology sensitivity by semantic category.
Convergent: both SAEs agree on anchored layer count ($\pm$2).
BTK$>$GS: fraction where BatchTopK shows stronger anchoring.}
\label{tab:per_token_category}
\vspace{-2mm}
\small
\begin{tabular}{@{}lrrrr@{}}
\toprule
\textbf{Category} & \textbf{Pairs} & \textbf{Tokens} & \textbf{Convergent} & \textbf{BTK$>$GS} \\
\midrule
Code/Math & 183 & 44 & \hlanchored{93\%} & 68\% \\
Content words & 202 & 91 & 48\% & 61\% \\
Function words & 85 & 21 & 29\% & 80\% \\
Numeric & 4 & 1 & 25\% & 100\% \\
\midrule
\textbf{All} & \textbf{474} & \textbf{157} & \textbf{62\%} & \textbf{67\%} \\
\bottomrule
\end{tabular}
\end{table}

Two patterns emerge.
First, \textbf{domain-specific tokens are SAE-robust}: code and math tokens (\texttt{operatorname}, \texttt{mathbf}, \texttt{createElement}) show 93\% convergence across SAE families, with matching anchor layer counts across 8--13 layers.
These tokens occupy a narrow, well-defined region in the training distribution, leaving little room for SAE methodology to alter their representation.
For example, the \texttt{operatorname} feature shows identical anchor counts (within $\pm$0 layers) across all 13 layers where both SAEs detect it.\footnote{\texttt{operatorname} single-token features: GemmaScope \href{https://neuronpedia.org/gemma-2-2b/3-gemmascope-res-16k/1853}{\texttt{L3\#1853}} (Neuronpedia: ``mathematical operators and functions''), BatchTopK \href{https://neuronpedia.org/gemma-2-2b/3-res-matryoshka-dc/940}{\texttt{L3\#940}} (``the string operatorname''); \texttt{mathbf} single-token features: GemmaScope \href{https://neuronpedia.org/gemma-2-2b/2-gemmascope-res-16k/212}{\texttt{L2\#212}} (``mathematical terminologies''), BatchTopK \href{https://neuronpedia.org/gemma-2-2b/2-res-matryoshka-dc/5067}{\texttt{L2\#5067}} (``linear algebra expressions'').
All identified via decoder-alignment detection (Section~\ref{sec:methods}).}

Second, \textbf{function words are SAE-dependent}: only 29\% convergence, with the highest BTK dominance (80\%).
The same function word can be causally necessary under one SAE but redundant under another.
For instance, ``al'' at L22 shows $\Delta$logit $= -0.38$ under GemmaScope but $-2.72$ under BatchTopK.\footnote{\texttt{al} single-token features: GemmaScope \href{https://neuronpedia.org/gemma-2-2b/22-gemmascope-res-16k/2521}{\texttt{L22\#2521}} (Neuronpedia: ``the word et''), BatchTopK \href{https://neuronpedia.org/gemma-2-2b/22-res-matryoshka-dc/736}{\texttt{L22\#736}} (``et al.\ citation abbreviation'').
Identified via decoder-alignment detection.} Function words are frequent enough that multiple SAE features can share their representation, making the allocation of causal importance sensitive to the competitive dynamics imposed by the activation function.

This category-dependent sensitivity suggests that SAE methodology comparisons should be stratified by token type: conclusions drawn from domain-specific tokens (where SAEs converge) may not generalize to function words (where they diverge).

\subsection{Magnitude-Matched Baseline}
\label{app:magnitude}

To test whether single-token feature ablation damage reflects activation magnitude rather than feature type, we compare each single-token feature against size-matched random controls evaluated on the same input positions: across Gemma-2-2B GemmaScope ($N\!=\!2{,}626$ pairs, 26 layers) and BatchTopK ($N\!=\!4{,}574$ pairs, 25 layers), single-token features cause more logit damage than controls under a pooled Wilcoxon test, $p < 10^{-94}$ and rank-biserial $r = 0.56$--$0.70$.
Stratifying by activation magnitude quartile, the effect holds even in the lowest quartile (Q1: $p < 0.0001$, $r = 0.27$--$0.43$) and strengthens with magnitude (Q4: $r = 0.69$--$0.83$)---this rules out the alternative explanation that single-token features are merely high-activation features whose ablation damage reflects magnitude rather than functional role.

\subsection{Control-Population Inertness by Configuration}
\label{app:control_inertness}

Every causal experiment pairs each single-token feature with five magnitude-matched random controls drawn from non-single-token features of the same SAE, measured at the same positions and target-token readout.
Table~\ref{tab:control_inertness} reports the pooled control population per full-depth configuration under the pipeline's stored recovery flag (ablated same-layer logit-lens rank within $2\times$ of baseline).
Control ablations are causally inert on the paired token readouts in every configuration and under both SAE families: recovery is at least 99.96\% and the median relative rank displacement is 1.000, so the anchored-versus-redundant family split does not appear in the non-single-token control population and is not a generic artifact of the ablation protocol.

\begin{table}[h]
\centering
\caption{Non-single-token control population per full-depth configuration: counts, median signed $\Delta\text{logit}$, recovery, and median relative rank displacement (1.000 = no effect). Single-token medians shown for reference.}
\label{tab:control_inertness}
\small
\resizebox{\columnwidth}{!}{%
\begin{tabular}{@{}lrrrrr@{}}
\toprule
\textbf{Configuration} & $n_{\text{ctrl}}$ & \textbf{Ctrl med.\ $\Delta$logit} & \textbf{Ctrl recovery} & \textbf{Ctrl rel.\ disp.} & \textbf{ST med.\ $\Delta$logit} \\
\midrule
Gemma-2-2B GemmaScope & 7{,}878 & $-0.00006$ & 100.0\% & 1.000 & $-0.0024$ \\
Gemma-2-2B BatchTopK & 22{,}870 & $-0.00006$ & 100.0\% & 1.000 & $-0.0050$ \\
Gemma-2-9B GemmaScope & 18{,}795 & $0.00000$ & 100.0\% & 1.000 & $-0.0055$ \\
Gemma-3-1B GemmaScope & 16{,}030 & $-0.00038$ & 100.0\% & 1.000 & $-0.0030$ \\
Gemma-3-1B BatchTopK & 11{,}325 & $-0.00043$ & 100.0\% & 1.000 & $-0.0270$ \\
Llama-3.1-8B LlamaScope & 14{,}360 & $-0.00120$ & 100.0\% & 1.000 & $-0.3744$ \\
DeepSeek-R1 LlamaScope & 18{,}245 & $0.00000$ & 100.0\% & 1.000 & $-0.0031$ \\
\bottomrule
\end{tabular}}
\end{table}

\subsection{Alignment-Matched Null Control}
\label{app:geonull}

Decoder-alignment detection selects features whose decoder vector aligns with the target token's input embedding, so the necessity effect could in principle be an artifact of that selection geometry.
The direct test is a null population matched on decoder--embedding cosine.

\paragraph{Selection procedure and matching tolerance.}
For each single-token feature we compute the cosine between every decoder vector in the SAE dictionary and the target token's input embedding, exclude all detected single-token features, and select controls within $\pm 0.02$ of the feature's own cosine; when fewer than five in-band candidates exist, we take the five nearest non-single-token features by cosine distance.
Exact matching is only partially constructible: the median single-token feature has 0--2 in-band candidates over the full 16k dictionary on Gemma-2-2B GemmaScope and 0 at all four analyzed Llama-3.1-8B layers, where single-token features at layer 1 have median cosine 0.65 against a far lower non-single-token maximum.
At single-token alignment levels, decoder--embedding alignment and single-token behavior nearly coincide as populations.
The achieved matching gap for the nearest-null controls is median $|\Delta\cos| = 0.18$ on GemmaScope and $0.51$ on LlamaScope.

\paragraph{In-band subset.}
Pooling the measured controls that fall within $\pm 0.02$ of their single-token feature's cosine across five Gemma-2-2B layers (315 in-band controls vs.\ 590 single-token features, same measurement protocol): control median signed $\Delta\text{logit}$ is $+0.00004$ versus $-0.0032$ for single-token features, one-sided Mann-Whitney $U$ $p = 6.1\times 10^{-20}$, rank-biserial 0.37.
Per layer the comparison is significant at 4 of 5 layers ($p = 3.1\times10^{-4}$, $3.5\times10^{-3}$, $2.8\times10^{-4}$, $1.4\times10^{-12}$); layer 6 has only 11 in-band controls and its point estimate reverses, so we report it as inconclusive.
Alignment also does not predict ablation damage within the control population: Spearman correlation between a control's cosine and its signed $\Delta\text{logit}$ is $-0.024$ ($p = 0.19$, $n = 2{,}950$ Gemma-2-2B control ablations).

\paragraph{Nearest-null comparison.}
Table~\ref{tab:geonull} reports single-token features against the five nearest-alignment non-single-token controls per target on Gemma-2-2B (five layers) and Llama-3.1-8B (four layers).
Necessity remains significant at eight of nine layer-configurations under the signed one-sided test, and the same eight survive BH correction within this nine-test family.
As a clustering sensitivity check, collapsing each single-token feature to a single paired comparison (Wilcoxon signed-rank on the feature's $\Delta\text{logit}$ minus the median of its five controls) leaves the same eight configurations significant.
The nearest-alignment controls carry a non-trivial effect of their own, most clearly on Llama-3.1-8B layer 1, so a geometric component of the necessity effect is detectable; it does not, however, account for the single-token effect, which exceeds the nearest available null in every configuration except Gemma-2-2B layer 6, where the single-token effect itself is smallest.

\begin{table}[h]
\centering
\caption{Single-token features vs.\ nearest-alignment non-single-token controls: mean signed $\Delta\text{logit}$ and one-sided Mann-Whitney $p$ per layer-configuration.}
\label{tab:geonull}
\small
\begin{tabular}{@{}llrrr@{}}
\toprule
\textbf{Family} & \textbf{Layer} & \textbf{ST mean} & \textbf{Ctrl mean} & $p$ \\
\midrule
GemmaScope & 1 & $-0.009$ & $-0.003$ & $0.024$ \\
GemmaScope & 6 & $-0.005$ & $-0.005$ & $0.34$ \\
GemmaScope & 12 & $-0.012$ & $+0.005$ & $6.7\times10^{-5}$ \\
GemmaScope & 18 & $-0.140$ & $+0.007$ & $6.6\times10^{-5}$ \\
GemmaScope & 24 & $-0.475$ & $-0.035$ & $7.7\times10^{-13}$ \\
LlamaScope & 1 & $-1.908$ & $-1.318$ & $1.2\times10^{-16}$ \\
LlamaScope & 8 & $-0.510$ & $-0.221$ & $0.011$ \\
LlamaScope & 16 & $-0.974$ & $-0.337$ & $1.1\times10^{-6}$ \\
LlamaScope & 24 & $-0.404$ & $-0.217$ & $0.013$ \\
\bottomrule
\end{tabular}
\end{table}

\end{document}